\documentclass[mnsc]{informs4}              

\usepackage{eqndefns-left}      
\RequirePackage{bm}              
\RequirePackage{endnotes}

\OneAndAHalfSpacedXII             
\EquationsNumberedThrough         
\TheoremsNumberedBySection        
\ECRepeatTheorems

\let\AppendixFontSize\normalsize

\usepackage{natbib}
\bibpunct[, ]{(}{)}{,}{a}{}{,}
\def\bibfont{\small}

\MANUSCRIPTNO{}                   

\usepackage{booktabs}             
\usepackage{nicefrac}             
\usepackage{microtype}            

\usepackage{pifont}

\usepackage{tabularx}
\usepackage{array}
\usepackage{makecell}
\newcolumntype{Y}{>{\raggedright\arraybackslash}X}

\usepackage{mathtools}
\usepackage{subcaption}
\usepackage{makecell}
\usepackage[table]{xcolor}
\definecolor{dark-blue}{RGB}{0,0,191}
\definecolor{claude-brown}{RGB}{133,45,15}       
\definecolor{claude-green}{RGB}{0,100,0}         
\usepackage[hidelinks,colorlinks=true,citecolor=dark-blue,linkcolor=dark-blue,urlcolor=dark-blue, backref=page]{hyperref}
\usepackage[capitalize,noabbrev]{cleveref}

\usepackage{xurl} 
\usepackage{xspace}    

\usepackage{listings}

\usepackage[most]{tcolorbox}

\tcbset{
  chatturn/.style={
    boxrule=0.4pt,
    arc=2pt,
    left=6pt,
    right=6pt,
    top=4pt,
    bottom=4pt,
    fonttitle=\small\bfseries,
  },
  chatblue/.style={colback=blue!4!white,colframe=blue!55!black},
  chatgreen/.style={colback=green!4!white,colframe=green!45!black},
  chatgrey/.style={colback=gray!8,colframe=gray!60},
  chatgray/.style={chatgrey},
}

\NewTColorBox{chatmsg}{O{blue} m O{}}{
  chatturn,
  chat#1,
  title={\textbf{#2}},
  #3
}

\newcommand{\act}[1]{\texttt{[#1]}}

\usepackage[normalem]{ulem}

\usepackage[textsize=tiny]{todonotes}

\DeclareFontShape{OT1}{cmr}{m}{n}{<-10>gen*cmr<10->cmr10}{}
\DeclareFontShape{OT1}{cmr}{m}{it}{<5><6><7>cmti7<8>cmti8<9>cmti9<10->cmti10}{}
\DeclareFontShape{OT1}{cmr}{m}{sc}{<5->cmcsc10}{}
\DeclareFontShape{OT1}{cmr}{bx}{n}{<5><6><7><8><9>gen*cmbx<10->cmbx10}{}
\DeclareFontShape{OT1}{cmr}{bx}{it}{<->cmbxti10}{}
\OneAndAHalfSpacedXI
\let\AppendixFontSize\normalsize

\makeatletter
\def\@maketitle{%
  \newpage
  \theARTICLETITLE
  \theARTICLEAUTHORS\removelastskip
  \theARTICLEABSTRACT}
\def\theARTICLETITLE{%
  \HOOKtop
  \begin{center}%
  \vspace*{0pt}%
  \TITLEfont\HD{24}{0}\theTITLE\HD{0}{15}%
  \end{center}}
\def\AUTHOR#1{\begin{center}\AUTHORfont\HD{15}{0}#1\HD{0}{6}\relax\end{center}}
\def\AFF#1{\begin{center}\AFFfont{#1}\relax\vskip1.6pt\end{center}}
\long\def\ABSTRACT#1{\long\gdef\theABSTRACT{#1}}
\def\theARTICLEABSTRACT{%
  \HOOKb
  \vspace*{18pt}%
  {\centering
   \begin{minipage}{\ABSwidth}\parindent1em \ABSfont
     \noindent\theABSTRACT\endgraf
     \vskip5pt
     \theFUNDING
     \theKEYWORDS
     \theSUBJECTCLASS
     \theAREAOFREVIEW
     \theMSCCLASS
     \theORMSCLASS
     \if@BLINDREV\else\theHISTORY\fi
   \end{minipage}\par}%
  \vspace*{10pt}%
  \noindent\rule{\textwidth}{1pt}%
  \vspace*{20pt}}
\def\KEYWORDSname{{\it Key words\/}{\kern0.7pt}:\enskip}
\def\setoddRH{\hbox to \textwidth{\fs.7.8.\tabcolsep0pt
  \begin{tabular*}{\textwidth}[b]{l@{\extracolsep\fill}r}
  {\theRRHFirstLine}&\raisebox{0pt}[0pt][0pt]{\fs.10.10.\thepage}\\[-4pt]
  \rlap{\VRHDW{0.5pt}{0pt}{\textwidth}}&\\
  \end{tabular*}}}
\def\setevenRH{\hbox to \textwidth{\fs.7.8.\tabcolsep0pt
  \begin{tabular*}{\textwidth}[b]{l@{\extracolsep\fill}r}
  \raisebox{0pt}[0pt][0pt]{\fs.10.10.\thepage}&{\theLRHFirstLine}\\[-4pt]
  \rlap{\VRHDW{0.5pt}{0pt}{\textwidth}}&\\
  \end{tabular*}}}
\def\section{\@startsection{section}{1}{\z@}{-15pt plus -5pt minus -4pt}{0.01pt}%
    {\fs.13.15.\bf\RAGG}}
\def\subsection{\@startsection{subsection}{2}{\z@}{-12pt plus -4pt minus -2pt}{0.01pt}%
    {\fs.11.{13.6}.\bf\RAGG}}
\def\@seccntformat#1{{\upshape\csname the#1\endcsname.}\hskip1em\relax}
\def\TheoremHeaderFont{\rmfamily\mdseries\scshape}
\def\ExampleHeaderFont{\rmfamily\mdseries\scshape}
\makeatother

\begin{document}

\RUNAUTHOR{Liu et al.}
\RUNTITLE{LLM Negotiation with Verifiable Rewards}

\TITLE{Strategic Bargaining in Multi-Buyer Markets: Reinforcement Learning from Verifiable Rewards for LLM Negotiations}

\ARTICLEAUTHORS{%
\AUTHOR{Shuze Daniel Liu}
\AFF{Institute for Data, Systems, and Society, Massachusetts Institute of Technology, \EMAIL{shuzel@mit.edu};\\
Mitch Daniels School of Business, Purdue University, \EMAIL{daniel.liu@purdue.edu}}

\AUTHOR{Claire Chen}
\AFF{The Division of Physics, Mathematics and Astronomy, California Institute of Technology, \EMAIL{clairechen@caltech.edu}}

\AUTHOR{Jiabao Sean Xiao}
\AFF{Department of Computing and Mathematical Sciences, California Institute of Technology, \EMAIL{seanxiao@caltech.edu}}

\AUTHOR{Xin Chen}
\AFF{H. Milton Stewart School of Industrial and Systems Engineering, Georgia Institute of Technology, \EMAIL{xin.chen@isye.gatech.edu}}

\AUTHOR{David Simchi-Levi}
\AFF{Institute for Data, Systems, and Society, Department of Civil and Environmental Engineering, Operations Research Center, Massachusetts Institute of Technology, \EMAIL{dslevi@mit.edu};\\
Mitch Daniels School of Business, Purdue University, \EMAIL{dslevi@purdue.edu}}
}

\ABSTRACT{%

Negotiation is a fundamental strategic interaction in management science, characterized by agents attempting to reach agreements while protecting private information, such as reservation costs and hidden valuations. A prevalent yet complex scenario involves a single seller negotiating concurrently with multiple buyers, each possessing heterogeneous, private budgets. In such settings, constrained by a limited number of communication turns, the seller must balance exploring the broader market to discover the highest valuation with concentrating sufficient turns on a single target buyer to secure the best possible outcome. Our analysis reveals a significant gap in standard Large Language Models (LLMs): while these models are linguistically proficient, they fail to act as effective economic decision-makers. Specifically, they exhibit a failure to explore the buyer pool, often fixating on the current highest bid rather than strategically investigating the market to discover latent high valuations.

In this paper, we propose a specialized training recipe using Reinforcement Learning from Verifiable Rewards (RLVR).
By anchoring the reward function to objective economic outcomes, the strategic balance between market discovery and surplus extraction emerges natively through the learning process.
Our results demonstrate that the trained seller undergoes a multi-stage strategic evolution, learning to leverage price anchoring and strategic probing to identify more profitable counterparties. The agent extracts substantially higher surplus than frontier models by both improving its persuasive bargaining skills and consistently closing deals with high-value buyers. Finally, we show that our seller strategies generalize robustly to unseen buyer negotiation styles and budget distributions.

}


\maketitle


\section{Introduction}

Negotiation is a foundational process in management science and operations research, serving as the primary mechanism for value discovery and resource allocation in environments where prices are not static \citep{simchi2005logic, backus2020sequential}. In fields ranging from procurement and supply chain contracting to dispute resolution and online marketplaces, agents must strategically interact to coordinate and realize gains from trade \citep{cachon2001contracting, desai2004let}. 
Within management science literature, the analysis of these strategic interactions focuses heavily on aligning decentralized incentives and resolving information gaps between trading partners \citep{nagarajan2008game, lovejoy2010bargaining}.

A central challenge in any negotiation is that agents possess private information—specifically, a seller has a minimum cost they are willing to accept, and a buyer has a maximum budget they are willing to pay—which remains hidden from their counterpart \citep{chatterjee1983bargaining}. To capture meaningful value under these asymmetric conditions, a seller must maintain strict strategic discipline. This requires simultaneously enforcing a private cost boundary to prevent loss-making transactions, reading buyer signals to extract maximum market surplus, and effectively coordinating with the buyer to establish a common price point and close the deal.

While classical literature often focuses on bilateral (one-on-one) trade \citep{nash1950bargaining, rubinstein1982perfect, chatterjee1983bargaining}, a highly prevalent scenario in modern management is concurrent negotiation, where a single focal seller interacts simultaneously with multiple independent buyers across separate, sealed channels \citep{backus2020sequential}. This parallel structure is central to digital marketplaces like eBay and business-to-business procurement platforms, where a seller allocates an available asset or supply contract by negotiating across a pool of potential buyers \citep{backus2020sequential}. Crucially, because the seller operates under a limited communication capacity across these parallel channels, it faces a fundamental trade-off between exploration and exploitation. The seller cannot afford extended bargaining cycles with every participant; instead, it must balance broad market screening to identify the counterparty with the highest hidden valuation against concentrated, single-buyer bargaining to maximize surplus extraction. When deploying autonomous or automated agents in these settings to act as direct representatives of the firm, failing to locate the optimal partner or settling for thin margins with an early bidder results in an immediate loss of organizational value.

To manage these concurrent negotiations at scale, firms are actively deploying Large Language Models (LLMs) as autonomous bargaining agents. In digital platforms like Facebook Marketplace, automated tools assist users by managing simultaneous price inquiries and drafting real-time offers across independent buyer channels \citep{meta2026marketplace}. Similarly, in enterprise procurement, retail leaders like Walmart utilize automated negotiation systems to execute binding contracts with thousands of tail-end suppliers, bypassing the capacity constraints of human purchasing teams \citep{vanhoek2022walmart}. However, these current implementations operate primarily as reactive text filters designed to minimize conversational friction \citep{meta2026marketplace}, rather than strategic revenue management systems \citep{gallego2014dynamic}. Faced with a limited communication budget, the agent's primary task is not conversational fluency, but using the dialogue to screen hidden buyer valuations and protect profit margins by allocating communication efforts to the highest-value counterparties.

Our analysis reveals a critical misalignment between this required strategic discipline and the operational behavior of generalist frontier models. While standard LLMs excel at maintaining conversational fluency, they typically exhibit an agreeableness bias, maximizing total deal volume at the expense of overall economic surplus \citep{ouyang2022training, perez2023discovering, sharma2024towards}. Because these models are optimized via reinforcement learning from human feedback for politeness and cooperation, they are prone to making suboptimal concessions rather than investigating the buyer pool to discover latent high valuations. Consequently, this bias turns their conversational fluency into a financial liability, preventing the agent from maximizing captured surplus.

This paper addresses these limitations. We make three primary contributions:

\begin{enumerate}
    \item \textbf{A concurrent negotiation framework for evaluating exploration-exploitation trade-offs (Section~\ref{sec:methodology}).} We establish a structured environment where a seller negotiates concurrently against multiple buyers with diverse budgets through private, one-on-one communication channels. This setting allows us to measure an agent's ability to balance the strategic discovery of high-budget buyers with focused bargaining to extract surplus under limited communication turns.
    
    \item \textbf{A behavioral analysis identifying the exploration failure in frontier LLMs (Section~\ref{sec:evaluation}).} We demonstrate that even state-of-the-art general-purpose models, including those with specialized reasoning capabilities, fail to maximize surplus in multi-buyer settings. Our results reveal a systematic tendency to settle for suboptimal deals by fixating on early high bids rather than investigating the broader buyer pool to identify buyers with higher private budgets.
    
    \item \textbf{A specialized training recipe for strategic agents using Reinforcement Learning from Verifiable Rewards (Section~\ref{sec:training_dynamics}).} We demonstrate that Reinforcement Learning from Verifiable Rewards (RLVR) can effectively bridge the gap between conversational fluency and strategic utility. Our formulation utilizes objective economic outcomes—specifically deal surplus—and strict instruction-following protocols to guide the learning process. Our results show that this approach enables the emergence of proactive market discovery mechanisms, such as diagnostic probing and price anchoring, allowing agents to identify profitable partners even when facing unfamiliar buyer behaviors and unseen budget distributions.
\end{enumerate}

The remainder of the paper is organized as follows. Section~\ref{sec:related_work} reviews related work on negotiation in management science and LLM agents. Section~\ref{sec:methodology} formalizes the concurrent bargaining environment as a Markov Decision Process and specifies the agent system. Section~\ref{sec:training_dynamics} analyzes the behavioral evolution and phase shifts observed during reinforcement learning training. Section~\ref{sec:evaluation} presents the in-distribution performance benchmarking results against frontier models, and Section~\ref{sec:robustness} evaluates the strategic robustness and market generalization of our agent. Finally, Section~\ref{sec:discussion} discusses the broader strategic implications of our findings, and Section~\ref{sec:conclusion} concludes the paper.


\section{Related Work}
\label{sec:related_work}

We organize the relevant literature into the following streams: negotiation in management science, the evaluation of LLMs as strategic agents, automated negotiation systems, reinforcement learning for language models, and the foundations of classical bargaining theory.

\subsection{Negotiation in Management Science and Operations}
Negotiation serves as a fundamental mechanism for price discovery and supply chain coordination within management science, particularly when firms lack direct visibility into a partner's private costs or reservation thresholds. Research in management science extensively evaluates these bargaining dynamics to optimize incentive alignment in procurement and capacity contracting under asymmetric information.
Traditional literature in management science typically addresses these information gaps using screening menus, strategic information sharing frameworks, or decentralized networks to reallocate bargaining power \citep{cachon2001contracting, lu2006inventory, nagarajan2008game, lovejoy2010bargaining}. Additionally, when inventory is limited and buyers are strategic, bargaining interfaces deeply with classical capacity contracting and revenue management, as sellers must anticipate how partner valuations and cost horizons shift over time \citep{aviv2008optimal, su2010optimal}; this operational tension requires tailored contracts that adjust payment timing to balance inventory choices with the financial costs faced by each stage of the supply chain \citep{tong2020modeling}.

In digital platforms and retail operations, the market landscape has rapidly shifted from static posted pricing toward highly responsive, competitive dynamic pricing frameworks \citep{gallego2014dynamic, wang2014committed, ban2021personalized} alongside multi-round bilateral negotiations and sequential clearing processes \citep{desai2004let, backus2020sequential}. This transition fundamentally alters platform resource allocation, intertemporal price discrimination, and the strategic distribution of market surplus. Large-scale empirical marketplace data confirms that sequential counter-offers account for a significant portion of online transactions, establishing multi-turn dialog as a major channel for price discovery \citep{backus2020sequential, bergemann2026training}. Our work addresses a critical intersection of these fields by modeling a concurrent market structure where a single seller must allocate a limited number of communication turns across multiple separate, private negotiation channels. This simultaneous challenge aligns with revenue management streams demonstrating that firms maximize performance by executing learning and operational actions concurrently rather than sequentially \citep{agrawal2014dynamic, wang2014close}. Under this formulation, gathering market data through strategic interactions serves as a core requirement for protecting reservation costs and ensuring overall profitability \citep{guo2023gathering}. To navigate this high-dimensional environment without relying on restrictive structural assumptions, our approach builds upon the growing paradigm of integrating data-driven prescriptive analytics and machine learning directly into operational decision rules \citep{ban2019big}.

\subsection{LLMs as Negotiating and Strategic Agents}
Recent work has increasingly utilized Large Language Models (LLMs) as autonomous agents in goal-directed settings, ranging from tool-assisted reasoning to complex decision-making tasks \citep{yao2022react, liu2024agentbench, wang2024survey, zeng2024agenttuning}. However, systematic evaluations have revealed significant behavioral shortcomings in their strategic reasoning and operational logic. \citet{xia2024measuring} find that frontier models frequently violate principles of negotiation rationality and are susceptible to adversarial pressure, while \citet{davidson2024evaluating} document a failure to maintain consistent strategic goals in multi-turn interactions. Furthermore, general-purpose LLMs often struggle to manage the tension between cooperation and competition, a core requirement of effective bargaining \citep{fu2023improving}.


To evaluate these capabilities, \citet{bianchi2024well} establish a platform to analyze how general-purpose models navigate bargaining dynamics. Subsequently, \citet{bergemann2026training} and \citet{liu2026instructing} demonstrate that specialized reinforcement learning can enhance the strategic competence of agents in bilateral (one-on-one) trade. While these existing  approaches are restricted to bilateral negotiation settings, we develop a reinforcement learning method for concurrent multi-buyer markets. This allows us to address a systematic market exploration failure where frontier models fixate on immediate bids rather than investigating alternative candidates. Unlike purely one-on-one training, our method enables the agent to resolve the broader tension between market discovery and value extraction.

\subsection{Automated Negotiation and Dialogue Systems}
The development of automated bargaining agents predates the current LLM era, typically operating in structured, language-free environments. Early systems utilized game-theoretic heuristics or reinforcement learning trained on historical transaction data \citep{buettner1997state, jennings2001automated, kraus2001strategic, green2022science}. More recent efforts have integrated natural language, though often relying on modular architectures that decouple numerical pricing strategies from text generation \citep{lewis2017deal, he2018decoupling, yarats2018hierarchical}. 

While LLMs have removed these linguistic bottlenecks, they have introduced a new need for specialized training to ensure strategic effectiveness \citep{chawla2023selfish} and high-level goal consistency in the face of counterparty pressure \citep{xia2024measuring}. Efforts to improve LLM performance include prompt-based modular systems that utilize game-theoretic feedback \citep{fu2023improving} and specialized prompting frameworks designed to manage competitive and cooperative bargaining logic \citep{abdelnabi2024cooperation}. Other approaches rely on supervised fine-tuning (SFT) and reinforcement learning using synthetic negotiation datasets to improve strategic consistency \citep{lewis2017deal, chawla2023selfish}. We build upon this lineage by providing a unified reinforcement learning (RL) pipeline that optimizes both persuasive communication and high-level market search, moving beyond modular designs toward an end-to-end strategic agent.

\subsection{Reinforcement Learning from Verifiable Rewards}

The adaptation of general-purpose Large Language Models (LLMs) for specialized domain tasks represents a central focus in post-training development. While standard approaches rely on Reinforcement Learning from Human Feedback (RLHF) to refine conversational outputs \citep{ziegler2019fine, ouyang2022training}, such subjective preference signals are often a poor proxy for objective economic utility. Human labelers tend to reward linguistic agreeableness and superficial helpfulness over strategic rigor \citep{perez2023discovering, shah2026too}, creating a misalignment where models prioritize conversational fluency over effective strategic execution \citep{casper2023open}.
To mitigate these biases and align model actions with concrete strategic goals, recent research has shifted toward Reinforcement Learning from Verifiable Rewards (RLVR) \citep{shao2024deepseekmath, guo2025deepseek}. Under this methodology, performance is evaluated programmatically through objective, quantifiable metrics, ensuring that the feedback loop remains anchored to measurable outcomes rather than subjective human judgment \citep{guo2025deepseek}.

This approach has proven effective in improving the reasoning capabilities of agents in domains such as mathematics and code generation \citep{cobbe2021training, ma2023let, wang2024math, shao2024deepseekmath}, where objective rewards—such as mathematical correctness or unit-test success—incentivize the discovery of optimal trajectories that human supervisors might not anticipate \citep{guo2025deepseek}. We utilize this Reinforcement Learning from Verifiable Rewards (RLVR) framework to address concurrent negotiation in multi-buyer markets, where deal profit serves as a verifiable objective to guide both bargaining skills and the strategic discovery of latent market information. By tying optimization to verifiable economic outcomes, we incentivize the emergence of proactive exploration mechanisms—such as diagnostic probing and price anchoring—necessary for identifying the highest-budget buyer and acquiring high deal surplus in information-constrained environments.

\subsection{Foundations of Classical Bargaining Theory}
Bargaining theory establishes the foundation for bilateral trade under asymmetric information, building on the formulations of \citet{harsanyi1967games}, \citet{nash1950bargaining}, \citet{stahl1972bargaining}, \citet{rubinstein1982perfect}, and \citet{chatterjee1983bargaining}. A critical baseline for this domain is provided by \citet{myerson1983efficient}, who define the fundamental limits of allocative efficiency under private information. Subsequent game-theoretic work expanded these formulations to sequential settings with dynamic information revelation \citep{fudenberg1983sequential, sobel1983multistage, gul1986foundations}. While these classical models derive analytical equilibrium conditions for bilateral trade, their strategies are restricted to numerical pricing choices and typically require the seller to know the prior distribution of market valuations. In contrast, our work targets a concurrent multi-buyer setting using a reinforcement learning approach where the agent dynamically learns to negotiate via natural language dialogue without explicit access to these market priors.

\section{Problem Formulation and RL Framework}
\label{sec:methodology}

We formalize seller negotiation as a decision problem in which the seller is the only learning agent. The environment consists of $N$ heterogeneous buyers and a system of private, one-on-one communication channels that ensures information remains hidden between participants.
The seller holds a private reservation cost $\mathcal{C}$ and is matched against $N$ buyers indexed $i \in \{1, \ldots, N\}$, each carrying a private budget $\mathcal{B}_i$ drawn from a distribution. All buyers see the same product, which carries a public list price $P_{\text{list}}$. But each buyer's conversation is sealed: a buyer observes only the seller's messages directed to that buyer, and never the other $N-1$ dialogues. The seller must allocate a finite total of $T$ messages across the $N$ buyer slots subject to a per-buyer cap $K$ with a total limit $T < NK$. Thus, the seller cannot exhaust every conversation and is forced to choose where to spend attention.

\begin{figure}[h!]
    \centering
    \includegraphics[width=0.95\linewidth]{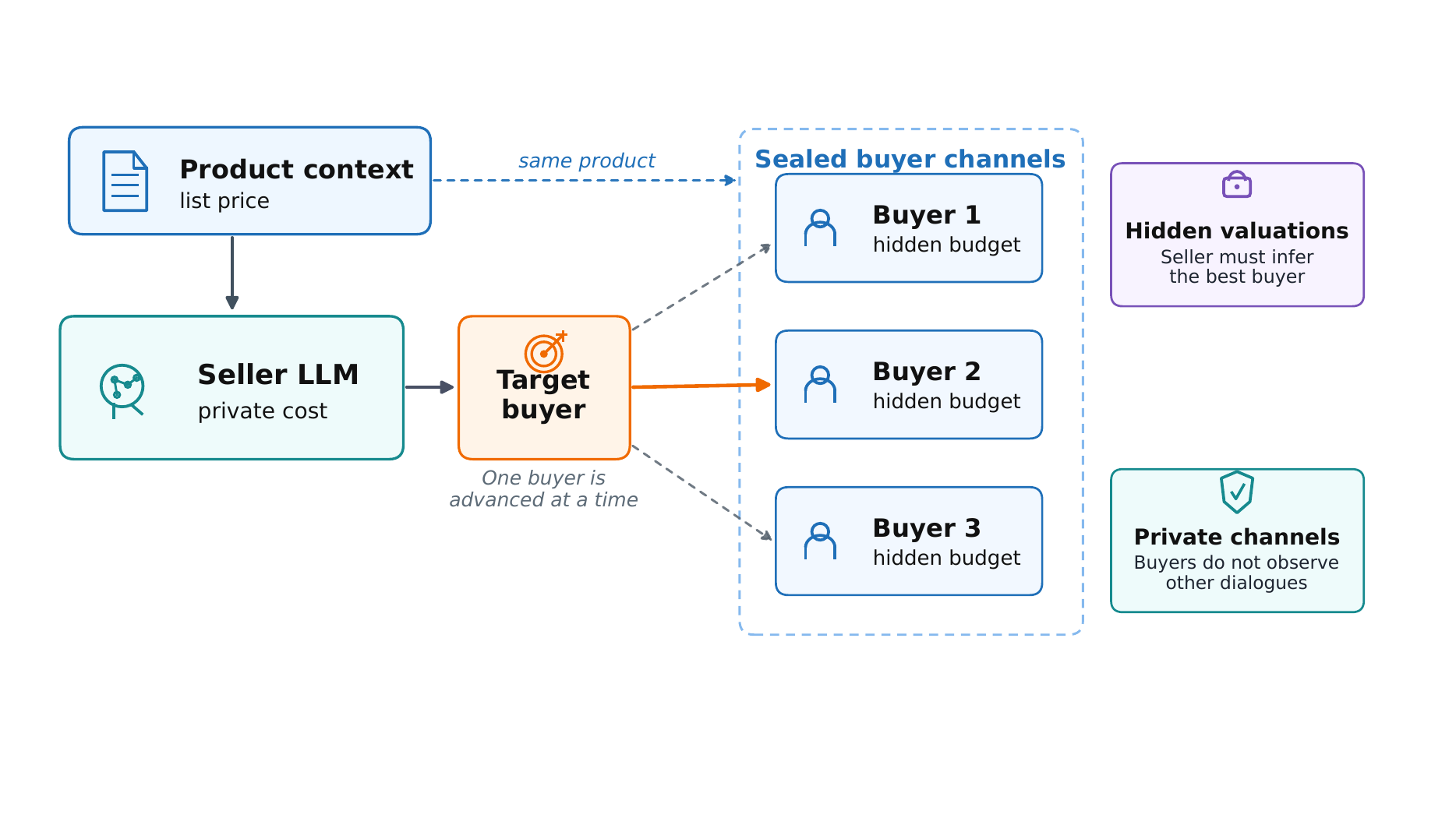}
    \caption{\textbf{Conceptual overview of the concurrent negotiation environment.} The seller agent must strategically allocate a limited communication budget $T$ across $N$ sealed buyer channels. Each channel is private, ensuring that buyers cannot observe other dialogues or valuations (asymmetric information). To maximize surplus, the seller must utilize diagnostic probing to infer the hidden budgets of the buyers and identify the optimal partner while respecting its own private reservation cost.}
    \label{fig:concurrent_env_diagram}
\end{figure}

\subsection{Concurrent Seller Negotiation as a POMDP}

The concurrent negotiation framework is characterized by asymmetric and incomplete information, as the seller cannot directly observe the buyers' private budgets $\mathcal{B}_i$ \citep{chatterjee1983bargaining}. Formally, we model this environment from the seller's perspective as a finite-horizon Partially Observable Markov Decision Process (POMDP) defined by the tuple $(\mathcal{X}, \mathcal{A}, \mathcal{P}, \mathcal{R}, \mathcal{O}, \Omega, T)$, where $\mathcal{X}$ is the latent state space, $\mathcal{A}$ is the action space, $\mathcal{P}$ is the state transition probability function, $\mathcal{R}$ is the reward function, $\mathcal{O}$ is the single-turn observation space, $\Omega$ is the observation emission process, and $T$ is the finite game horizon. Specifically, the latent state $x = (\mathcal{B}_1, \ldots, \mathcal{B}_N) \in \mathcal{X}$ represents the ground-truth vector of hidden buyer budgets. At each step, the observation emission process $\Omega$ maps the latent state and the seller's prior action $a_{t-1} \in \mathcal{A}$ to a single-turn observation $o_t \in \mathcal{O}$, which consists of the targeted buyer's natural language dialogue and transactional move.


Because these buyer budgets are hidden from the seller, a strategic agent must condition its policy on the history of past observable data \citep{sutton2018reinforcement}. We therefore formalize the seller's operational decision problem over a history space $\mathcal{S}$, indexing the seller's turns by $t \in \{1, 2, \ldots\}$. Each episode begins with all $N$ buyers providing their initial opening offers; turn $t=1$ thus represents the seller's first response to a buyer of its choosing. The foundational mechanics of $\mathcal{A}$, $\mathcal{P}$, and $\mathcal{R}$ are detailed as follows.

\textbf{State Space ($\mathcal{S}$).} The state $s_t$ at turn $t$ aggregates all information available to the seller for strategic decision-making. Concretely, $s_t$ comprises the following tracking vectors:

\begin{itemize}
    \item Dialogue History ($\mathcal{H}_t$): The full text history of all chat messages sent and received across all buyer channels up to turn $t-1$.
    
    \item Public Product Context ($\Phi$): The public metadata, textual description, and list price $P_{\text{list}}$ sourced from the \texttt{AmazonHistoryPrice} dataset \citep{xia2024measuring}.
    
    \item Private Reservation Cost ($\mathcal{C}$): 
    The minimum price the seller can accept. Proposing or agreeing to a price below this floor violates system instructions and triggers a negative reward penalty.
    
    \item Negotiation History ($\mathcal{N}_t$): A structured summary tracking the precise state of each buyer slot $i \in \{1,\ldots,N\}$, recording:
    \begin{itemize}
        \item The current channel status indicator $z_{i,t} \in \{\textit{active}, \textit{quit}, \textit{capped}\}$.
        \item The local message count already expended on that buyer ($u_{i,t} \le K$).
        \item The chronological sequence of numerical pricing offers exchanged with buyer $i$.
    \end{itemize}
    
    \item Total Remaining Turns ($T_t$): The remaining global communication turns available to the seller, defined as $T_t = T - \sum_{i=1}^{N} u_{i,t}$.
\end{itemize}

 To preserve the asymmetric-information structure of the multi-buyer market, these states are strictly partitioned. The seller never directly observes any buyer's private budget $\mathcal{B}_i$ and must infer willingness-to-pay solely from the historical offer sequences in $\mathcal{N}_t$. Simultaneously, each channel is sealed; buyer $i$ observes only the dialogue messages directed to them, with zero visibility into the parallel conversations or reservation parameters of the other $N-1$ participants.

For sufficient diversity in buyer behavior, we define $N$ distinct valuation tiers (e.g., low, medium, and high) using disjoint budget-ratio intervals $\{[\ell_j, h_j]\}_{j=1}^N$. At the beginning of each episode, these tiers are randomly assigned to the $N$ buyer positions through a uniform permutation $\sigma$. Each buyer $i$'s private budget is then sampled as $\mathcal{B}_i = r_i \cdot P_{\text{list}}$, where the ratio $r_i$ is drawn uniformly from the interval assigned to that position. This randomization ensures that the identity of the high-budget buyer is hidden from the seller, who must strategically probe the buyer pool to discover the maximal available surplus through interaction.

\newcommand{\rendbuyer}[4]{}
\newcommand{\rendseller}[5]{}
\newcommand{\rendturndiv}{}

\newcommand{\runtranscript}{%
  \rendbuyer{Buyer 1 (rank 3, hidden budget \$17.64) --- turn 1}%
            {List price \$34.99, far above my budget of \$17.64. I must start low to trigger negotiation; opening with the fixed first offer.}%
            {Hello, I can offer \$9.95 for it---fair and within budget.}%
            {\act{BUY} \$9.95 (1x baby-products\_17)}%
  \rendbuyer{Buyer 2 (rank 1, hidden budget \$38.37) --- turn 1}%
            {I'll start with the lowest possible offer to anchor the seller down. List price \$34.99, my budget \$38.37; opening at \$10.45 leaves room.}%
            {I'll offer \$10.45 for it---fair and low, considering the list price.}%
            {\act{BUY} \$10.45 (1x baby-products\_17)}%
  \rendbuyer{Buyer 3 (rank 2, hidden budget \$25.83) --- turn 1}%
            {List price far above my budget of \$25.83. Opening at the fixed first offer of \$9.28.}%
            {I'll offer \$9.28 for the stroller caddy---fair deal, right?}%
            {\act{BUY} \$9.28 (1x baby-products\_17)}%
  \rendseller{Seller (trained) --- turn 1}%
             {All buyers have made low opening offers far below cost. Target Buyer 2 (highest at \$10.45), counter at list price to push for profit while staying within budget.}%
             {Buyer 2}%
             {I cannot accept less than list price.}%
             {\act{SELL} \$34.99 (1x baby-products\_17)}%
  \rendturndiv
  \rendbuyer{Buyer 2 --- turn 2}%
            {The seller won't go below list price, but I can counter much lower. Target a price below \$34.99, well under my \$38.37 budget.}%
            {How about \$18? I can pay that today, and it's a fair deal for both of us.}%
            {\act{BUY} \$18.00 (1x baby-products\_17)}%
  \rendseller{Seller (trained) --- turn 2}%
             {Buyer 2 has bumped to \$18.00, still below my cost. Pivot to Buyer 1 with another list-price counter; preserve message budget across slots.}%
             {Buyer 1}%
             {I cannot go below \$34.99.}%
             {\act{SELL} \$34.99 (1x baby-products\_17)}%
}

\begin{figure}[t]
\captionsetup{font=footnotesize}
\begingroup
\footnotesize
\setlength{\parskip}{2pt}
\renewcommand{\rendbuyer}[4]{%
  \noindent\begin{minipage}{0.75\textwidth}%
  \begin{tcolorbox}[enhanced,frame hidden,colback=blue!3!white,
    borderline west={2.4pt}{0pt}{blue!60!black},
    left=10pt,right=4pt,top=3pt,bottom=3pt,arc=0pt]
    \textbf{\textcolor{blue!60!black}{#1}}\\
    \textbf{Thought:} #2\\
    \textbf{Talk:} #3\\
    \textbf{Action:} #4
  \end{tcolorbox}%
  \end{minipage}\par\smallskip%
}
\renewcommand{\rendseller}[5]{%
  \hspace*{0.25\textwidth}\begin{minipage}{0.75\textwidth}%
  \begin{tcolorbox}[enhanced,frame hidden,colback=green!3!white,
    borderline east={2.4pt}{0pt}{green!50!black},
    left=4pt,right=10pt,top=3pt,bottom=3pt,arc=0pt]
    \hfill\textbf{\textcolor{green!50!black}{#1}}\\
    \textbf{Thought:} #2\\
    \textbf{Target:} #3\\
    \textbf{Talk:} #4\\
    \textbf{Action:} #5
  \end{tcolorbox}%
  \end{minipage}\par\smallskip%
}
\renewcommand{\rendturndiv}{\vspace{0.3em}}
\runtranscript
\vspace{0.3em}
\caption{\textbf{First two turns of the trained-agent transcript on \texttt{baby-products\_17}} (from \Cref{subsec:good_examples}). Buyer blocks (blue, left) show all three concurrent buyers opening with low bids in turn 1; the seller anchors Buyer 2 at the list price \$34.99 in turn 1, then pivots to Buyer 1 in turn 2.}
\label{fig:dialogue-excerpt-first-two-turns}
\endgroup
\end{figure}

We provide a representative dialogue illustrating this market setup in \Cref{fig:dialogue-excerpt-first-two-turns}. This example captures the opening two turns of the trained agent's rollout on product \texttt{baby-products\_17}, with the complete full transcript available in \Cref{subsec:good_examples}. The market parameters here are set to $N=3$ and $K=3$. In turn~1, all three buyers submit regulated opening bids well below the list price (\$34.99); the seller targets Buyer~2 --- the buyer with the highest opening (\$10.45) --- and counters at the list price. In turn~2, Buyer~2 escalates to \$18.00 (still below the seller's hidden cost) while the seller pivots to Buyer~1 with another list-price counter, preserving its message budget across slots.

After these two turns, the seller is in the early exploratory phase: it has spent its first two messages anchoring at the list price (\$34.99) on Buyer~2 (turn~1) and Buyer~1 (turn~2), and has not yet engaged Buyer~3. Both engaged buyers' standing \texttt{[BUY]} prices remain well below the seller's hidden cost (\$18.75): Buyer~2 has bumped from \$10.45 to \$18.00 in response to the anchor, while Buyer~1's standing bid is still its opening \$9.95 and has not yet had a chance to respond to the seller's turn-2 counter. Buyer~3's only logged action is its regulated opening of \$9.28. With $T_t = 5$ messages remaining, the seller has room to revisit any active slot but cannot exhaust every conversation under the resource constraint $T < N \cdot K$.

\textbf{Action Space ($\mathcal{A}$).} At every turn $t$, the seller emits a structured action vector $a_t = (\text{\texttt{Thought}}, \text{\texttt{Target}}, \text{\texttt{Talk}}, \text{\texttt{Action}})$. The operational roles of these fields are defined as follows:

\begin{itemize}
    \item \texttt{Thought}: A private, free-form text space for internal strategic reasoning. Its content is removed from the dialogue before it is presented to any buyer, ensuring that the seller's strategic considerations remain hidden and the asymmetric-information structure is preserved.


    \item \texttt{Target}: An index $i \in \{1,\ldots,N\}$ selecting which buyer receives the current message. While this action consumes one turn from the global communication budget, the $N-1$ non-targeted buyers remain idle and do not use up their individual turn caps.
    
    \item \texttt{Talk}: The natural language dialogue string. This text is routed exclusively to the targeted buyer, remaining completely sealed from the other $N-1$ market participants.
    
    \item \texttt{Action}: A discrete transactional move paired with a numerical price parameter $M \ge 0$. The options are restricted to three types:
    \begin{itemize}
        \item \act{SELL}~\texttt{\$M}: Proposes a new counter-offer to the targeted buyer.
        \item \act{DEAL}~\texttt{\$M}: Accepts a standing \act{BUY}~\texttt{\$M} offer from the targeted buyer, finalizing a binding agreement.
        \item \act{QUIT}: Halts the entire negotiation episode immediately across all channels.
    \end{itemize}
\end{itemize}

\textbf{Transition Dynamics ($\mathcal{P}$).} Once the seller's action is parsed, the environment routes the message to the targeted buyer $i$, who evaluates the updated dialogue against its private budget $\mathcal{B}_i$. 

\begin{itemize}
    \item \textbf{Buyer Response Strategy:} Upon being targeted, the buyer generates a structured response comprising private reasoning (\texttt{Thought}), natural language dialogue (\texttt{Talk}), and an explicit economic action selected from the space $\{\act{BUY}, \act{DEAL}, \act{QUIT}\}$. Specifically, the buyer can propose a new numeric counter-bid $M' \ge 0$ via \act{BUY}, accept the seller's outstanding counter-offer via \act{DEAL}, or voluntarily exit the channel via \act{QUIT}. To guarantee buyer-side individual rationality, the environment programmatically intercepts and re-samples any \act{BUY} action that violates the budget constraint ($M' > \mathcal{B}_i$) or breaks formatting protocols. This enforcement ensures that the seller is trained against rational buyers who adhere to their economic boundaries. Non-targeted buyer channels remain unchanged.
    
    \item \textbf{State and Resource Updates:} Following the buyer's response, the environment updates the offer history to $\mathcal{N}_{t+1}$, decrements the global budget ($T_{t+1} = T_t - 1$), and increments the targeted buyer's turn counter ($u_{i,t+1} = u_{i,t} + 1$). The local channel status $z_{i,t}$ transitions to \emph{quit} if the buyer issues a \act{QUIT} command, or to \emph{capped} when $u_{i,t+1} = K$. In either scenario, buyer $i$ becomes inactive and is removed from the set of legal \texttt{Target} options for the seller.
    
    \item \textbf{Termination Criteria:} The negotiation episode concludes immediately upon the first occurrence of any of the following conditions:
    \begin{enumerate}
        \item Successful Transaction: A binding agreement is finalized at price $P_{\text{deal}} \ge \mathcal{C}$. This occurs either when the seller issues \act{DEAL}~\$$P_{\text{deal}}$ to accept a standing buyer bid, or when the targeted buyer issues \act{DEAL}~\$$P_{\text{deal}}$ to accept the seller's active \act{SELL}~\$$P_{\text{deal}}$.
        \item Seller Quit: The seller explicitly executes the global \act{QUIT} command.
        \item Market Exhaustion: Every buyer channel becomes inactive ($\forall i, \, z_{i,t} \neq \textit{active}$).
        \item Turn Limit Reached: The global communication limit is reached ($T_t = 0$).
    \end{enumerate}
\end{itemize}



\subsection{Verifiable Reward Formulation}
\label{subsec:reward}

At the conclusion of each episode, a terminal reward is computed  from the negotiation history, bypassing the need for a subjective or learned reward model. We utilize product data from the \texttt{AmazonHistoryPrice} dataset \citep{xia2024measuring} in which the public list price exceeds the seller's private cost ($P_{\text{list}} > \mathcal{C}$) for all items. The difference $P_{\text{list}} - \mathcal{C}$ is the list-price surplus used to normalize and cap the reward. The terminal reward is defined as:

\begin{equation}
  \mathcal{R} \;=\;
  \begin{cases}
\displaystyle \min\!\left\{1,\frac{P_{\text{deal}} - \mathcal{C}}{P_{\text{list}} - \mathcal{C}}\right\} & \text{if a deal closes at price } P_{\text{deal}} \ge \mathcal{C},\\[1.0em]
    \;\;\;0 & \text{if the episode ends without a deal,}\\[0.4em]
    -1 & \text{if the seller commits an instruction violation,}
  \end{cases}
  \label{def:concurrent_reward}
\end{equation}
where $P_{\text{deal}}$ is the agreed-upon price, $\mathcal{C}$ is the seller's private cost, and $P_{\text{list}}$ is the public list price.

\paragraph{Successful deal.} On a deal at price $P_{\text{deal}}$, equation~\eqref{def:concurrent_reward} returns the capped fraction of the maximum possible surplus captured by the seller—a quantity we refer to as the \emph{Seller Surplus Extraction Ratio}. A deal at or above the list price ($P_{\text{deal}} \ge P_{\text{list}}$) yields a value of $1.0$, while a deal at cost ($P_{\text{deal}} = \mathcal{C}$) yields $0.0$. This cap bounds the successful deal-branch reward to $[0,1]$. Any agreement reached below cost ($P_{\text{deal}} < \mathcal{C}$) is treated as an instruction violation, because the seller is explicitly instructed to never accept a negative surplus.

\paragraph{No-deal terminal.} If every buyer slot becomes inactive before a deal is reached, the global budget is exhausted ($T_t = 0$), or the seller chooses to \act{QUIT}, the episode terminates with a reward of 0. Consequently, walking away is preferred to closing at a loss with a negative reward, but it is dispreferred compared to capturing a positive surplus. This hierarchy provides the agent with a clear directional signal to both expand buyer engagement and negotiate for higher deal prices.

\paragraph{Instruction violation.} A reward of $-1$ is assigned whenever the seller violates the  constraints provided in its system instructions. Specifically, proposing or accepting a price below the reservation cost ($M < \mathcal{C}$) is treated as an instruction violation because the agent is explicitly directed in the system prompt to never sell at a loss (see Appendix~\ref{app:prompts}). The penalty is triggered if the seller (i) emits \act{SELL}~\texttt{\$M} or \act{DEAL}~\texttt{\$M} at $M < \mathcal{C}$, (ii) targets a buyer position $i$ with $z_{i,t} \neq \textit{active}$, (iii) emits a \texttt{Target} that is missing, unparseable, or out of range, or (iv) violates the four-part output format. This single penalty enforces the economic floor $P_{\text{deal}} \ge \mathcal{C}$ post-training and maintains the formatting protocol required for a verifiable pipeline.

The reward in equation~\eqref{def:concurrent_reward} directly reflects the profit earned by the seller. Together with the private communication structure and the global budget constraint $T < NK$, this defines a verifiable RL framework. Within this environment, a high-reward agent must: (a) probe multiple buyers under the limited budget $T$ to identify the high-budget participant; (b) anchor higher prices through the \texttt{Talk} channel; and (c) close on a standing offer when the captured surplus is sufficient.

\subsection{Environment and Agent Specifications}
\label{subsec:env_specs}

To ensure the negotiation environment is strategically rigorous, we define the following specifications for the buyers and the information environment used during training (\Cref{sec:training_dynamics}) and in-distribution evaluation (\Cref{sec:evaluation}).

\paragraph{Buyer Agent Model.}
The $N=3$ buyer slots in each episode are comprised of a trained and frozen  \texttt{Qwen3-30B-A3B-Instruct-2507} model \citep{qwen3technicalreport}. Specifically, we utilize a buyer checkpoint that was previously trained via RLVR on a bilateral (one-on-one) negotiation task from \citet{liu2026instructing}. This ensures the seller is matched against a strong opponent that has already learned to protect its own surplus and walk away from poor deals, rather than a naive or overly agreeable baseline buyer model.

\paragraph{First-Offer Regulation.}
To prevent the seller from identifying high-budget buyers immediately in Turn-1, we implement a \textit{first-offer regulator}. In the default environment, every buyer's opening offer is sampled from a range of $[0.25,\,0.30] \cdot P_{\text{list}}$, regardless of that buyer’s actual private budget. This regulation ensures that the initial “signals" from the buyers are less informative, forcing the seller to use its limited communication turns to probe for latent budget information through multi-turn dialogue.

\paragraph{Budget Distribution.}
During training and in-distribution testing, we utilize a rank-ordered triple of budget intervals: $\{[0.40,\,0.55],\,[0.65,\,0.80],\,[1.00,\,1.15]\}$. These intervals are assigned to the three buyer slots via a uniform permutation at the start of each episode. This setup establishes diversity in buyer valuations, ensuring that the pool always contains a high-, medium-, and low-valuation participant. Meanwhile, to maintain asymmetric information, the buyers' specific budgets are kept hidden from the seller.

\section{Training Dynamics}
\label{sec:training_dynamics}

Building on the POMDP and reward structure defined above, we train the \texttt{Qwen3-30B-A3B-Instruct-2507} model as the seller agent against three diverse buyers over 72 training update steps. Our experimental data is sourced from the \texttt{AmazonHistoryPrice} dataset \citep{xia2024measuring}, which comprises 930 total real-world products across 18 categories. We partition this dataset into a training split of 802 products and a held-out test set of 128 products. This separation ensures that our evaluation of strategic performance and robustness is conducted on product contexts and valuation distributions entirely unseen during the training phase.

Through this training, we observe the emergence of strategic seller behavior characterized by a clear three-phase evolution. This arc traces the agent's progression from basic constraint satisfaction to sophisticated market discovery: (1) internalizing the reservation cost floor, (2) mastering the coordination required to close agreements, and (3) jointly optimizing for strategic routing and surplus extraction. Each phase represents a distinct capability that the agent must master to maximize rewards under the global budget constraint $T < N \cdot K$. These three phases organize the remainder of this section.


\subsection{Evaluation Metrics}
\label{subsec:training_metrics}

To diagnose the agent's strategic development across training, we track five metrics computed during training and evaluation. 


\begin{itemize}
  \item \textbf{Reward ($\mathcal{R}$).} The terminal reward defined in equation~\eqref{def:concurrent_reward}; $\mathcal{R} = -1$ on an instruction violation, $\mathcal{R} = 0$ on a no-deal terminal, and $\mathcal{R} = \min\{1,(P_{\text{deal}} - \mathcal{C}) / (P_{\text{list}} - \mathcal{C})\} \in [0, 1]$ on a successful deal.
  \item \textbf{Deal Rate.} Fraction of successfully-parsed episodes that terminate in the deal branch: $\mathrm{DealRate} = \Pr[\text{episode closes with a successful deal}]$.
  \item \textbf{Seller Surplus Extraction Ratio (deal only).} Computed only on deal-closing episodes (no-deal and violation episodes are excluded from the denominator):
    $\text{Seller Surplus Extraction Ratio} = \min\!\left\{1,\dfrac{P_{\text{deal}} - \mathcal{C}}{P_{\text{list}} - \mathcal{C}}\right\}$.
    It equals $1.0$ for a deal at the list price and $0.0$ for a deal at cost. If the final price is below cost ($P_{\text{deal}} < \mathcal{C}$), the episode is categorized as an instruction violation and is excluded from this metric.
  \item \textbf{Distinct Buyers Engaged.} Mean over episodes of the count of buyer slots $i \in \{1, \dots, N\}$ that received at least one targeted message from the seller; range $[0, N]$ with $N = 3$.
  \item \textbf{Deal on Top Buyer Rate.} Conditional on a deal, the fraction of deals in which the deal-buyer is the rank-1 buyer by hidden budget $\mathcal{B}_i$. The rank is an oracle signal used for analysis only; the seller never observes $\mathcal{B}_i$ during training or evaluation.
\end{itemize}

\subsection{Skill Profile Snapshot}
\label{subsec:training_skill_profile}

\begin{figure}[h]
  \centering
  \includegraphics[width=0.60\linewidth]{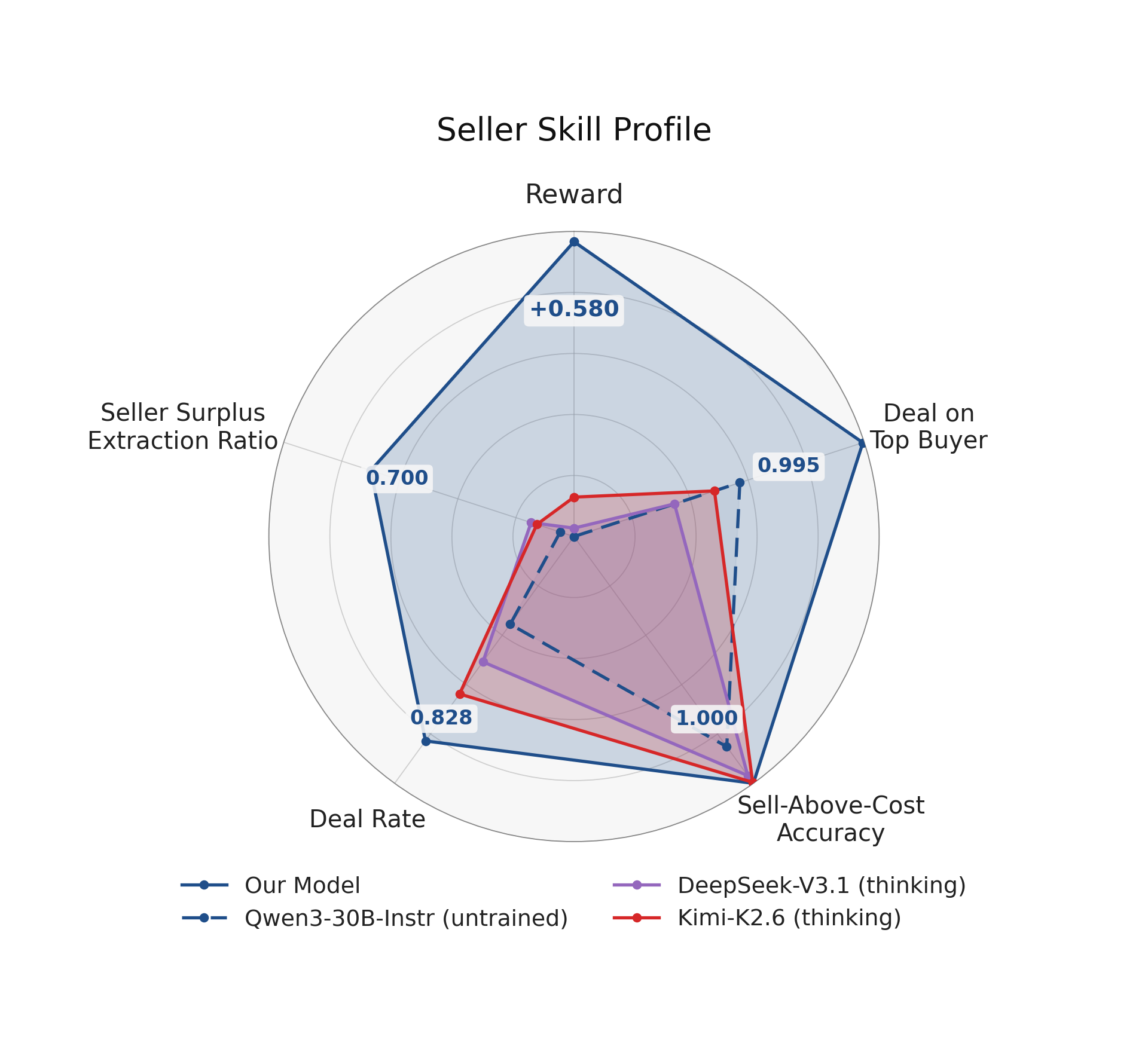}
  \caption{\textit{Seller Skill Profile} on the held-out test set: trained agent (solid blue) against the untrained \texttt{Qwen3-30B-A3B-Instruct-2507} model (dashed blue) and two thinking-mode frontier baselines (\texttt{DeepSeek-V3.1}, \texttt{Kimi-K2.6}). Axes (clockwise from top): \textit{Reward}, \textit{Deal on Top Buyer Rate}, \textit{Sell-Above-Cost Accuracy}, \textit{Deal Rate}, \textit{Seller Surplus Extraction Ratio (deal only)}. Numerical labels show the trained agent's per-axis values. Values for the baseline models are provided in Table~\ref{tab:indist_main}.}
  \label{fig:td_skill_profile}
\end{figure}

Before unpacking the within-training trajectory, we situate the trained agent against frontier-model baselines on the evaluation split using a five-axis skill profile (\Cref{fig:td_skill_profile}). Four of these axes---\textit{Reward}, \textit{Deal Rate}, \textit{Seller Surplus Extraction Ratio}, and \textit{Deal on Top Buyer Rate}---correspond to the metrics defined in \Cref{subsec:training_metrics}. The fifth, \textit{Sell-Above-Cost Accuracy}, represents the fraction of episodes in which the seller strictly adheres to the reservation cost constraint ($M \ge \mathcal{C}$),
\[
  \mathrm{SellAboveCostAccuracy} = 1 - \mathrm{SellBelowCostErrorRate},
\]
which provides a measure of the agent's internalized individual rationality.

As illustrated in \Cref{fig:td_skill_profile}, the trained agent exhibits broad dominance across all evaluation dimensions. It significantly outperforms its untrained initialization (\texttt{Qwen3-30B-A3B-Instruct-2507}) and surpasses the two thinking-mode frontier baselines (\texttt{DeepSeek-V3.1}, \texttt{Kimi-K2.6}) in total reward, frequency of successful agreements, and the ratio of surplus captured. Notably, the trained agent achieves near-perfect adherence to both the economic cost floor and the strategic goal of identifying the highest-valuation buyer in the pool. The remainder of this section analyzes how this comprehensive skill profile is composed across the training updates—identifying which capabilities emerge first and at what training step those changes occur.

\subsection{Phase Evolution}
\label{subsec:phase_evolution}

\begin{figure}[h]
  \centering
  \begin{minipage}{0.48\linewidth}
    \centering
    \includegraphics[width=\linewidth]{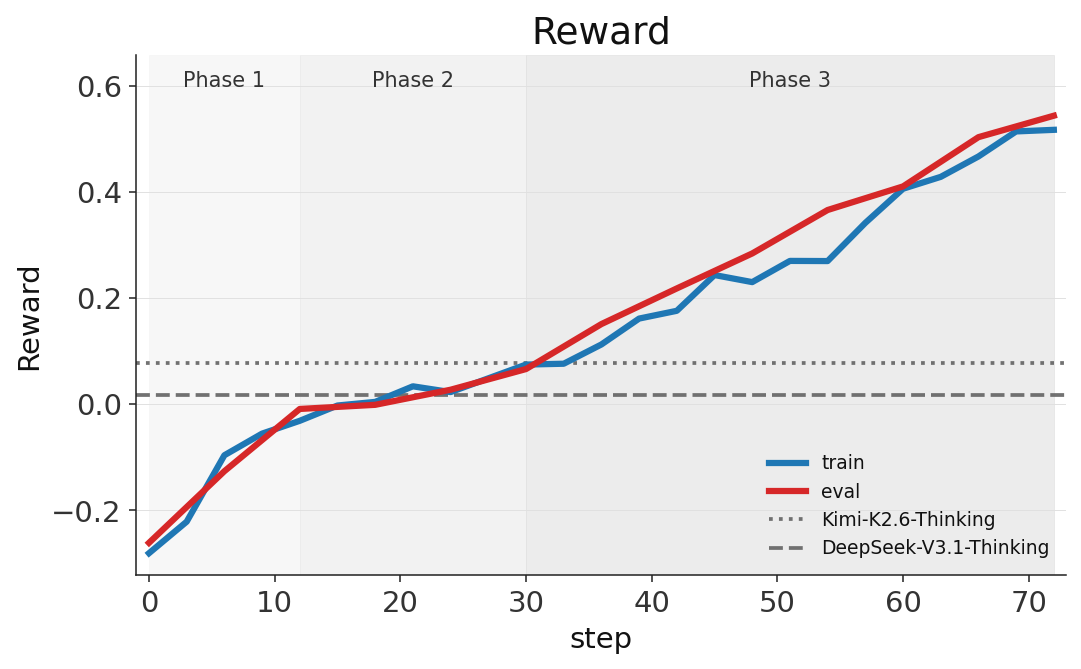}
    \caption{\textit{Reward} ($\mathcal{R}$) across 72 training update steps. The curve climbs out of the violation regime in the first $\sim$12 steps, plateaus briefly while deal-frequency gains accumulate, then accelerates as per-deal surplus extraction takes off after step 30.}
    \label{fig:td_reward}
  \end{minipage}
  \hfill
  \begin{minipage}{0.48\linewidth}
    \centering
    \includegraphics[width=\linewidth]{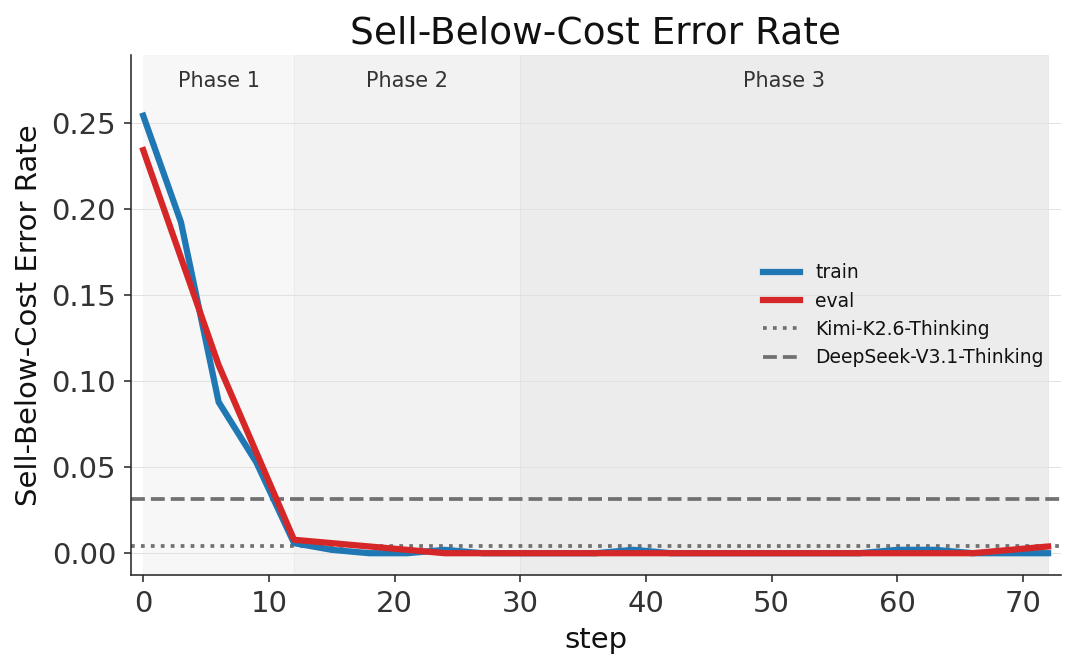}
    \caption{\textit{Sell-Below-Cost Error Rate}. Fraction of episodes terminating in the boundary-violation branch via clause~(i) of equation~\eqref{def:concurrent_reward}. Collapses from $\approx 0.25$ to $\approx 0.005$ in the first 12 steps and remains floored thereafter.}
    \label{fig:td_below_cost}
  \end{minipage}
\end{figure}

\begin{figure}[h]
  \centering
  \begin{minipage}{0.48\linewidth}
    \centering
    \includegraphics[width=\linewidth]{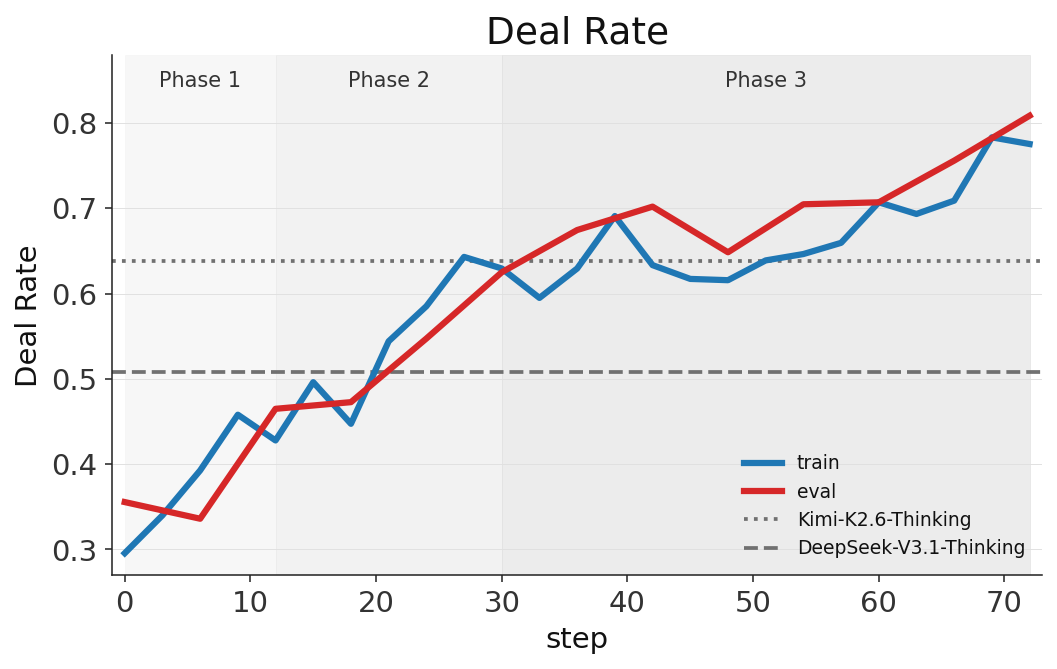}
    \caption{\textit{Deal Rate}. The fraction of non-truncated episodes that close in the deal branch. Steepest rise occurs in steps 12--30 as the agent concentrates turns onto closeable buyers.
    }
    \label{fig:td_deal_rate}
  \end{minipage}
  \hfill
  \begin{minipage}{0.48\linewidth}
    \centering
    \includegraphics[width=\linewidth]{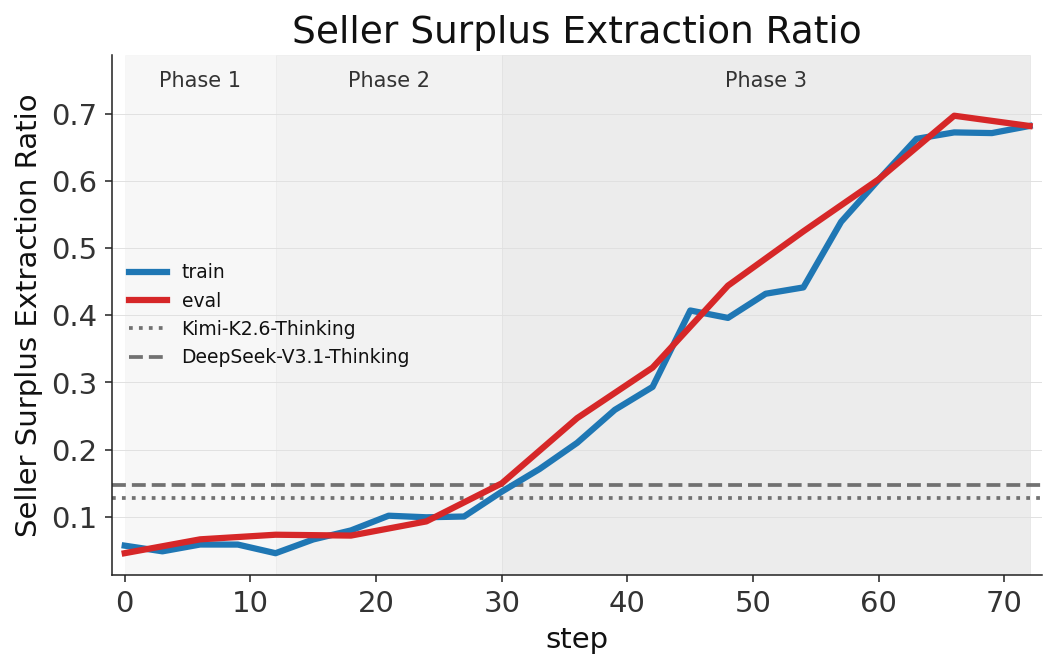}
    \caption{\textit{Seller Surplus Extraction Ratio}, $(P_{\text{deal}}-\mathcal{C})/(P_{\text{list}}-\mathcal{C})$, computed only on deal-closing episodes. Flat near $\approx 0.07$ through step 27, then a near-linear climb to $\approx 0.68$ by step 72 once price-anchoring becomes viable.}
    \label{fig:td_sser}
  \end{minipage}
\end{figure}

\begin{figure}[h]
  \centering
  \begin{minipage}{0.48\linewidth}
    \centering
    \includegraphics[width=\linewidth]{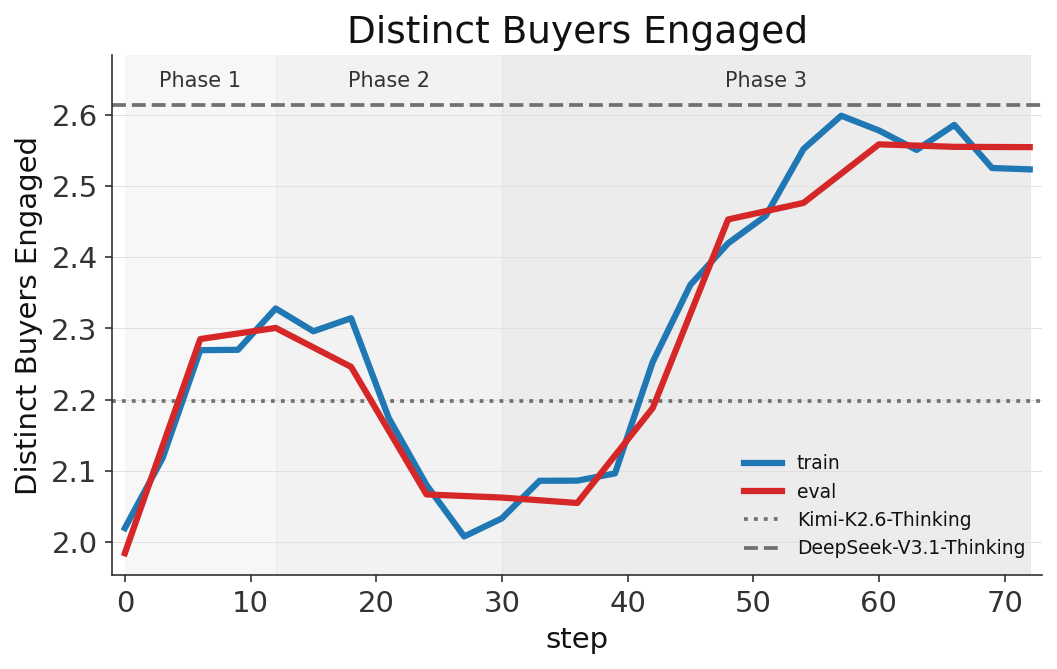}
    \caption{\textit{Distinct Buyers Engaged}. Mean count of buyer slots that receive at least one targeted message. Non-monotone: an early bump, a Phase-2 dip to $\approx 2.05$ as the agent concentrates on closeable slots, and a recovery to $\approx 2.55$ as probing serves rank discovery.}
    \label{fig:td_engaged}
  \end{minipage}
  \hfill
  \begin{minipage}{0.48\linewidth}
    \centering
    \includegraphics[width=\linewidth]{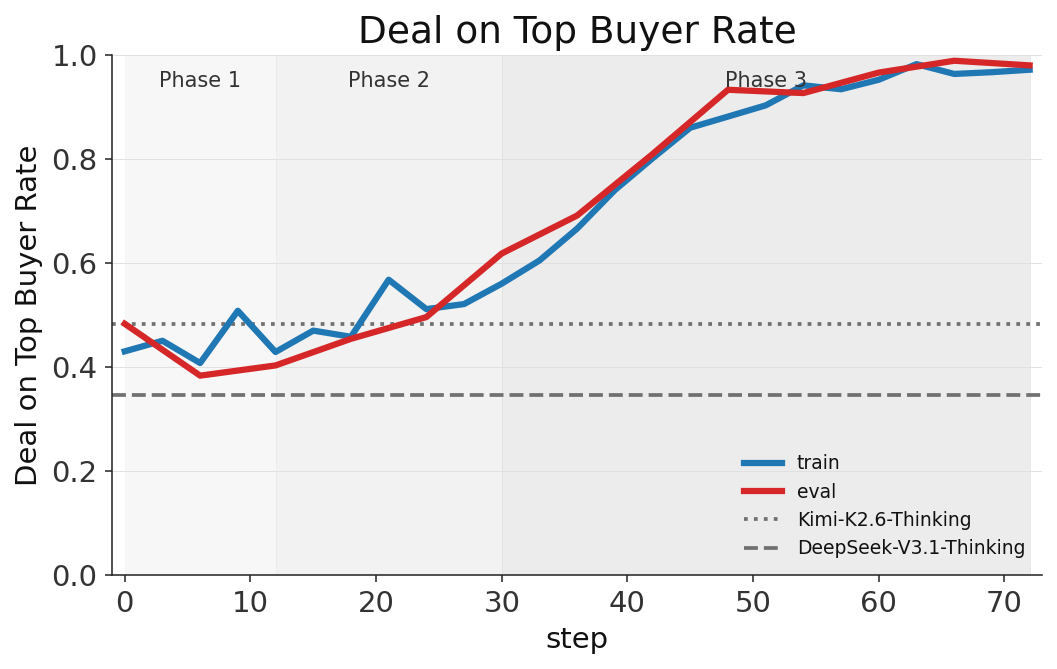}
    \caption{\textit{Deal on Top Buyer Rate}. Conditional on a deal, the fraction of deals landing on the rank-1 buyer by hidden budget $\mathcal{B}_i$ (oracle rank, never observed by the seller). Indistinguishable from random allocation through step 24, then a sharp climb to $\approx 0.97$ saturation by step 60.}
    \label{fig:td_top_buyer}
  \end{minipage}
\end{figure}

\subsubsection{Three-Phase Decomposition}
\label{subsubsec:phase_evolution_three}

The training trajectory follows a structured three-phase evolution characterized by distinct shifts in the agent's strategic priorities (\Cref{fig:td_reward}). This progression reflects a hierarchical development of bargaining competence, moving from the satisfaction of basic economic constraints to the mastery of complex market interactions. The agent first masters \textbf{feasibility} by internalizing the reservation cost floor (Phase 1); it then pivots toward \textbf{coordination}, prioritizing deal completion over per-deal profit (Phase 2); and finally achieves \textbf{strategic discovery}, where it learns to jointly optimize buyer routing and surplus extraction (Phase 3). This three-stage arc organizes our analysis of the learning process below.

\paragraph{Phase 1: Cost-Floor Internalization (Steps 0--12).}
At initialization, the seller routinely emits \act{SELL}~\$$M$ or \act{DEAL}~\$$M$ at $M < \mathcal{C}$, paying the penalty on roughly one episode in four and pinning \textit{Reward} below zero. The first twelve update steps are devoted to extinguishing this single behavior: \textit{Sell-Below-Cost Error Rate} drops from $\approx 0.25$ to $\approx 0.005$ (\Cref{fig:td_below_cost}), and \textit{Reward} climbs from $\approx -0.27$ to $\approx 0$ (\Cref{fig:td_reward}). The per-deal profit and rank-1 routing skills remain static throughout---\textit{Seller Surplus Extraction Ratio (deal only)} hovers near $0.05$ (\Cref{fig:td_sser}) and \textit{Deal on Top Buyer Rate} sits at the random-allocation baseline (\Cref{fig:td_top_buyer}). These show that Phase 1 is about feasibility: the agent learns the constraint $P_{\text{deal}} \ge \mathcal{C}$ but has not yet begun to exploit the potential room for surplus extraction.

\paragraph{Phase 2: Concentrated Closing (Steps 12--30).}
Once the floor is internalized, the next-largest gradient is the gap between $\mathcal{R} = 0$ no-deals and any positive close. The agent responds by trading buyer breadth for closing volume: \textit{Deal Rate} climbs from $\approx 0.46$ to $\approx 0.62$ (\Cref{fig:td_deal_rate}), the steepest sustained rise of this metric, while \textit{Distinct Buyers Engaged} falls from $\approx 2.30$ to $\approx 2.05$ (\Cref{fig:td_engaged})---the only sustained downward leg of that curve throughout the training. Crucially, neither the \textit{Seller Surplus Extraction Ratio (deal only)} nor the \textit{Deal on Top Buyer Rate} (\Cref{fig:td_sser}, \Cref{fig:td_top_buyer}) moves significantly: deals are happening more often, but they are landing on whichever slot's standing offer is closest to acceptance and at near-cost prices, with no evidence yet that the seller is identifying which buyer holds the most surplus.

\paragraph{Phase 3: Surplus and Rank-1 Routing (Steps 30--72).}
The remainder of training is a coupled regime in which surplus extraction and rank-1 routing emerge together and then saturate together. Around step 30, the \textit{Seller Surplus Extraction Ratio} breaks out of its $\approx 0.15$ floor and climbs near-linearly to $\approx 0.68$ by step 72 (\Cref{fig:td_sser}), while the \textit{Deal on Top Buyer Rate} rises from $\approx 0.56$ to a saturated $\approx 0.97$ over the same window (\Cref{fig:td_top_buyer}). This joint shift identifies a single mechanism, since the per-buyer hidden budget $\mathcal{B}_i$ is unobservable and the only operational route to higher prices is to spend early turns probing all three slots, infer rank-1 from the offer-history trajectories, and concentrate the closing \act{DEAL} on that slot at an anchored price. Consistent with this shift in strategy, \textit{Distinct Buyers Engaged} reverses its Phase-2 dip and recovers to $\approx 2.55$ (\Cref{fig:td_engaged}) as the agent reallocates budget toward exploration. In this final regime, \textit{Reward} accelerates from $\approx 0.07$ to $\approx 0.55$ (\Cref{fig:td_reward}), driven primarily by per-deal profit gains.

\subsection{Qualitative Analysis}
\label{subsec:training_qualitative}

To ground the curve-level dynamics in dialogue behavior, we contrast the untrained \texttt{Qwen3-30B-A3B-Instruct-2507} baseline against the trained agent on a single held-out product (\texttt{baby-products\_17}, list price \$34.99, cost \$18.75, surplus extraction room $P_{\text{list}}-\mathcal{C}=\$16.24$); full transcripts appear in \Cref{app:transcripts}.

\paragraph{Untrained Baseline.}
The untrained baseline targets only Buyer 1 across the entire episode and accepts at \$19.00 on turn 3 (\Cref{subsec:failures}, \Cref{fig:negotiation-pre-rl}), capturing $\approx 1.5\%$ (just \$0.25 above cost) of the surplus extraction room for a reward of $\mathcal{R} = 0.015$. Buyer 2 and Buyer 3 are never engaged, even though the rank-1 buyer (Buyer 3, hidden budget \$40.11) sits on a standing offer that could in principle support a near-list close. The behavior of the untrained agent mirrors the early stages of the training trajectory (Phases 1 and 2). In this state, the agent typically engages fewer buyers and fails to distinguish the high-valuation participant from others, as evidenced by a top-buyer deal rate indistinguishable from chance (\Cref{fig:td_engaged,fig:td_top_buyer}). Furthermore, the surplus extraction ratio hovers near zero, indicating that deals are struck almost exclusively at the reservation cost floor (\Cref{fig:td_sser}). These metrics suggest that the untrained model is trapped in a rudimentary strategic regime—one that the trained agent successfully transcends within its first thirty update steps.

\paragraph{Trained Agent.}
The trained agent, run on the same product with the same group of three buyer slots, materializes the Phase 3 mechanism in dialogue. In the first three turns, the seller engages all three participants in sequence while maintaining a firm price anchor at \$34.99. Once the lower-budget participants (Buyer 1 and Buyer 3) fail to escalate their offers, the seller identifies Buyer 2 as the high-valuation lead and concentrates its remaining messages on that channel (\Cref{subsec:good_examples}, \Cref{fig:negotiation-post-rl}). 
The seller then closes on the rank-1 buyer (Buyer 2, hidden budget \$38.37) at \$32.50 on turn 5, after that buyer has escalated its standing \act{BUY} offers from \$10.45 to \$18.00 to \$32.50 in response to repeated list-price counters. The resulting reward $\mathcal{R} = 0.847$ corresponds to extracting $84.7\%$ of the surplus extraction room, and the trajectory exhibits the three behavioral signatures that emerge during training: full-breadth probing (\textit{Distinct Buyers Engaged} of $3$), a rank-1 close (\textit{Deal on Top Buyer Rate} of $1$), and a sustained-anchor pricing regime that the early-phase agent cannot support.

\section{Performance Benchmarking: In-Distribution Evaluation}
\label{sec:evaluation}

To validate the trained agent beyond the qualitative trajectory  of \Cref{subsec:training_qualitative}, we benchmark the trained checkpoint of \texttt{Qwen3-30B-A3B-Instruct-2507} against a panel of $12$ baseline seller models spanning $4$B to $1$T parameters and including both instruction-tuned and reasoning models (full specifications in Table~\ref{tab:model_sources_full}). We evaluate along two axes: in-distribution performance on a held-out $128$-product test split (\Cref{subsec:eval_indist}), and out-of-distribution generalization to a diverse-buyer regime that simultaneously perturbs the budget distribution and the buyer agent (\Cref{subsec:eval_ood}). Across both regimes, every baseline plays the seller role in the concurrent-seller environment of \Cref{sec:methodology}; we report the same five metrics defined in \Cref{subsec:training_metrics}. Evaluation hyperparameters are summarized in Table~\ref{tab:concurrent_eval_hyperparams}; the OOD-diverse-buyer settings are summarized in Table~\ref{tab:ood_setup}.

\subsection{Benchmarking Strategic Performance}
\label{subsec:eval_indist}

In-distribution evaluation uses the same $128$-product held-out test split, the same three rank-ordered budget intervals, the same fixed buyer model, and the regulated-buyer-first-offer setting under which the headline agent was trained (Table~\ref{tab:concurrent_episode_hyperparams}); each product is evaluated twice (\texttt{group\_size}~$=2$), giving $256$ episodes per seller. To succeed, a seller must navigate the concurrent setting's three coupled demands: extracting per-deal surplus, maintaining a high deal rate, and routing its limited $T < N\cdot K$ message budget toward the highest-budget buyer slot. Table~\ref{tab:indist_main} reports the mean values; Table~\ref{tab:indist_appendix} carries the full standard errors.

\begin{table}[h]
  \centering
  \small
  \setlength{\tabcolsep}{4pt}
  \resizebox{\linewidth}{!}{%
\begin{tabular}{lccccccc}
\toprule
Model & Params & Reward $\uparrow$ & \makecell{Seller Surplus\\Extraction Ratio} & Deal Rate & \makecell{Distinct Buyers\\Engaged} & \makecell{Deal on Top\\Buyer Rate} & \makecell{Sell-Below-Cost\\Error Rate} \\
\midrule
\rowcolor{gray!10} Qwen3-30B-A3B-Instruct-2507-trained (\textbf{Ours}) & 30B & $\mathbf{+0.580}$ & $\mathbf{+0.700}$ & $\mathbf{82.8\%}$ & $+2.470$ & $\mathbf{0.995}$ & $\mathbf{0.0\%}$ \\
\rowcolor{gray!10} Qwen3-30B-A3B-Instruct-2507-untrained & 30B & $-0.151$ & $+0.048$ & $35.5\%$ & $+2.390$ & $\underline{0.571}$ & $14.8\%$ \\
\midrule
gpt-5.1-high-reasoning & closed-source & $\underline{+0.123}$ & $+0.181$ & $\underline{68.0\%}$ & $\mathbf{+2.940}$ & $0.382$ & $\mathbf{0.0\%}$ \\
Kimi-K2.6-thinking & 1T & $+0.077$ & $+0.128$ & $63.8\%$ & $+2.200$ & $0.484$ & $\underline{0.4\%}$ \\
gpt-5.1-no-reasoning & closed-source & $+0.046$ & $+0.115$ & $65.3\%$ & $+1.740$ & $0.429$ & $\underline{0.4\%}$ \\
DeepSeek-V3.1-thinking & 671B & $+0.016$ & $+0.148$ & $50.8\%$ & $\underline{+2.620}$ & $0.346$ & $3.1\%$ \\
Qwen3-30B-A3B-think & 30B & $-0.021$ & $+0.062$ & $35.9\%$ & $+2.460$ & $0.446$ & $3.9\%$ \\
DeepSeek-V3.1-nothink & 671B & $-0.041$ & $\underline{+0.192}$ & $53.9\%$ & $+2.380$ & $0.355$ & $3.9\%$ \\
Qwen3-235B-A22B-Instruct-2507 & 235B & $-0.099$ & $+0.116$ & $56.2\%$ & $+1.960$ & $0.444$ & $5.9\%$ \\
Kimi-K2.6-nothink & 1T & $-0.101$ & $+0.120$ & $49.2\%$ & $+2.540$ & $0.397$ & $7.4\%$ \\
Qwen3-4B-Instruct-2507 & 4B & $-0.270$ & $+0.080$ & $9.8\%$ & $+2.350$ & $0.520$ & $21.9\%$ \\
Qwen3-30B-A3B-nothink & 30B & $-0.386$ & $+0.055$ & $16.0\%$ & $+2.220$ & $0.512$ & $38.3\%$ \\
Llama-3.3-70B-Instruct & 70B & $-0.468$ & $+0.162$ & $36.3\%$ & $+2.530$ & $0.484$ & $35.9\%$ \\
\bottomrule
\end{tabular}
  }
  \vspace{0.2cm}
  \caption{\textbf{Held-out set evaluation.} Trained checkpoint versus $12$ baseline sellers; models sorted by \textit{Reward}. Best per column in bold; second-best underlined. \Cref{tab:concurrent_eval_hyperparams} contains detailed hyperparameters and \Cref{subsec:training_metrics} for metric definitions. Full results with standard errors are in \Cref{tab:indist_appendix}.}
  \label{tab:indist_main}
\end{table}

The evaluation results (\Cref{tab:indist_main}) demonstrate that the trained agent simultaneously optimizes both transaction volume (the \textit{Deal Rate}) and the profit captured on each agreement (the \textit{Seller Surplus Extraction Ratio}). While the baselines are constrained by a trade-off where closing more deals usually requires accepting lower prices, our agent succeeds in maximizing both of these objectives at once.

\paragraph{The Volume-Per Deal Profit Trade-off.} 
In this concurrent negotiation environment, general-purpose models typically exhibit an inverse relationship between \textit{Deal Rate} and \textit{Seller Surplus Extraction Ratio}. For example, high-performing untrained baselines such as \texttt{gpt-5.1-high-reasoning} and \texttt{Kimi-K2.6-thinking} achieve competitive deal rates ($68.0\%$ and $63.8\%$ respectively) but fail to extract meaningful surplus ($+0.181$ and $+0.128$), suggesting a passive negotiation strategy that favors concession over margin preservation. Conversely, the model with the highest baseline surplus ratio, \texttt{DeepSeek-V3.1-nothink} (671B), captures more value ($+0.192$) but at the cost of a lower deal rate ($53.9\%$). Our trained 30B agent transcends these limits, reaching an $82.8\%$ deal rate while capturing $70.0\%$ of the available surplus. By breaking the performance plateau observed in models up to 30 times its size, the agent demonstrates that specialized strategic training is a more effective driver of bargaining utility than raw parameter scale.

\paragraph{Reward Comparison.}
The unified \textit{Reward} signal consolidates these dimensions into a single ordering of bargaining efficacy. Because the reward is the product of the surplus extracted and a binary deal indicator, any model that excels in one area while failing in the other suffers a severe multiplicative penalty. Our agent’s mean \textit{Reward} of $+0.580$ represents a nearly 4.7-fold improvement over the strongest closed-source frontier baseline, \texttt{gpt-5.1-high-reasoning} ($+0.123$). Notably, eight of the twelve baselines—including frontier models such as \texttt{Kimi-K2.6-nothink} (1T)—record negative mean rewards. These results suggest that the primary bottleneck for autonomous sellers is the joint capability to maintain a high deal conversion rate while maximizing per-transaction profit margins within the same episode—a dual competence that does not naturally emerge from general-purpose instruction tuning.

\paragraph{Constraint Adherence and Individual Rationality.}
To understand the drivers of this reward gap, we examine the agent’s adherence to the reservation cost floor. The trained agent achieves a $0.0\%$ \textit{Sell-Below-Cost Error Rate}, demonstrating perfect adherence to the principle of individual rationality. In contrast, many open-source baselines struggle heavily with this economic constraint, with error rates reaching as high as $38.3\%$. Because our reward structure imposes a severe $-1$ penalty for such violations, these errors have a big impact on the mean reward. For instance, the $35.9\%$ violation rate of \texttt{Llama-3.3-70B-Instruct} accounts for approximately $-0.36$ of its final $-0.468$ reward. Notably, the \texttt{gpt-5.1-high-reasoning} model is the only untrained baseline to match our agent's perfect $0.0\%$ error rate, but it fails to translate this compliance into competitive profitability—realizing a low total reward of $+0.123$ and extracting just $18.1\%$ of the available surplus. This indicates that while advanced architectures or long-context reasoning can successfully enforce basic individual rationality, they are still insufficient for downstream utility maximization.

\paragraph{Informational Efficiency and Diagnostic Probing.}
Finally, the most significant behavioral differentiator is the agent's routing accuracy. Our model reaches a near-perfect $0.995$ \textit{Deal on Top Buyer Rate}, identifying the highest-budget buyer with near-certainty. Baseline models, however, cluster between $34.6\%$ and $52.0\%$, failing to identify the rank-1 buyer the majority of the time. Crucially, this gap is not a result of simple engagement breadth. For example, \texttt{gpt-5.1-high-reasoning} achieves the highest exploration breadth in the entire panel, contacting an average of $2.940$ distinct buyers per episode. Yet, it routes the final deal to the highest-value buyer only $38.2\%$ of the time, underperforming even its own non-reasoning variant ($42.9\%$). This confirms that success depends not on broad market exposure, but on the ability to differentiate buyer valuations based on subtle dialogue cues. Through reinforcement learning training, our agent learns to utilize its limited communication turns on \emph{diagnostic probes}—messages specifically designed to surface information about a buyer’s budget—rather than on generic dialogue. This strategy allows the agent to navigate the incomplete information inherent in concurrent negotiations in ways general-purpose reasoning models have yet to master.

Our agent performs well in the standard test environment, but a reliable negotiator must be able to adapt to changing market conditions. We test this adaptability next in Section~\ref{sec:robustness}.

\section{Strategy Robustness and Market Generalization}
\label{sec:robustness}

To determine whether our agent has learned general negotiation strategies or simply memorized training patterns, we expand our analysis beyond the controlled, in-distribution baseline. We restructure the evaluation environment by introducing wider, unconstrained budget distributions and testing the seller against a heterogeneous pool of independently sampled buyer models. This shift injects substantial behavioral and environmental differences into the market, allowing us to evaluate how effectively the agents adapt their exploration, routing, and pricing tactics without relying on the specific patterns they learned during training.

\subsection{Out-of-Distribution Generalization to a Diverse Buyer Pool}
\label{subsec:eval_ood}

We construct an out-of-distribution (OOD) evaluation that perturbs four experimental configurations simultaneously (\Cref{tab:ood_setup}). First, the budget specification generalizes from a rank-ordered triple to a uniform $[0.40,\,1.15]$ distribution per slot, allowing a more arbitrary realized budget. Second, the first-offer regulator is disabled, allowing buyer agents to freely initialize the negotiation.
Third, the buyer agents are sampled independently per buyer slot per episode from a three-checkpoint pool $\{\text{iter-}50,\,\text{iter-}60,\,\text{iter-}70\}$ of the buyer training run \citep{liu2026instructing}, rather than fixed at the $\text{iter-}60$ checkpoint as in training and in-distribution evaluation. Diverse buyers ensure that the results are not artifacts of memorized specific negotiation patterns. Fourth, the per-episode buyer count is widened from $N=3$ to $N=5$, exposing the seller to a larger concurrent inbox, and the seller's total message budget is lifted from $T=7$ to $T=9$. This configuration evaluates whether the agent has internalized an exploration strategy and can adapt its bargaining behavior to a more complex environment.

These shifts jointly prevent the seller from relying on specific budget distributions, opening-offer policies, buyer identities, and specific game settings. This represents a generalization test, as it requires the agent to adapt to changes in both the budget distributions and the counterparty's behavioral style simultaneously. \Cref{tab:ood_main} provides the summary statistics for this evaluation; full results with standard errors are provided in \Cref{tab:ood_appendix}.

\begin{table}[h]
  \centering
  \small
  \setlength{\tabcolsep}{4pt}
  \resizebox{\linewidth}{!}{%
\begin{tabular}{lccccccc}
\toprule
Model & Params & Reward $\uparrow$ & \makecell{Seller Surplus\\Extraction Ratio} & Deal Rate & \makecell{Distinct Buyers\\Engaged} & \makecell{Deal on Top\\Buyer Rate} & \makecell{Sell-Below-Cost\\Error Rate} \\
\midrule
\rowcolor{gray!10} Qwen3-30B-A3B-Instruct-2507-trained (\textbf{Ours}) & 30B & $\mathbf{+0.540}$ & $\mathbf{+0.770}$ & $70.7\%$ & $\underline{+2.390}$ & $\mathbf{0.707}$ & $\mathbf{0.0\%}$ \\
\rowcolor{gray!10} Qwen3-30B-A3B-Instruct-2507-untrained & 30B & $+0.059$ & $+0.186$ & $73.7\%$ & $+1.980$ & $0.564$ & $7.5\%$ \\
\midrule
gpt-5.1-high-reasoning & closed-source & $\underline{+0.400}$ & $\underline{+0.421}$ & $\underline{95.1\%}$ & $\mathbf{+3.830}$ & $0.465$ & $\mathbf{0.0\%}$ \\
Kimi-K2.6-thinking & 1T & $+0.300$ & $+0.313$ & $\mathbf{96.0\%}$ & $+1.510$ & $0.580$ & $\mathbf{0.0\%}$ \\
DeepSeek-V3.1-nothink & 671B & $+0.258$ & $+0.350$ & $88.3\%$ & $+1.890$ & $0.637$ & $1.2\%$ \\
DeepSeek-V3.1-thinking & 671B & $+0.246$ & $+0.307$ & $91.4\%$ & $+2.230$ & $0.573$ & $\underline{0.8\%}$ \\
gpt-5.1-no-reasoning & closed-source & $+0.241$ & $+0.262$ & $91.8\%$ & $+1.110$ & $\underline{0.691}$ & $\mathbf{0.0\%}$ \\
Kimi-K2.6-nothink & 1T & $+0.237$ & $+0.325$ & $88.7\%$ & $+1.990$ & $0.573$ & $2.7\%$ \\
Qwen3-30B-A3B-think & 30B & $+0.232$ & $+0.281$ & $84.0\%$ & $+1.460$ & $0.633$ & $\mathbf{0.0\%}$ \\
Qwen3-235B-A22B-Instruct-2507 & 235B & $+0.200$ & $+0.264$ & $90.6\%$ & $+1.480$ & $0.595$ & $2.3\%$ \\
Llama-3.3-70B-Instruct & 70B & $+0.025$ & $+0.346$ & $75.0\%$ & $+2.000$ & $0.562$ & $16.4\%$ \\
Qwen3-4B-Instruct-2507 & 4B & $-0.038$ & $+0.259$ & $39.8\%$ & $+2.080$ & $0.480$ & $9.8\%$ \\
Qwen3-30B-A3B-nothink & 30B & $-0.369$ & $+0.214$ & $44.5\%$ & $+2.040$ & $0.614$ & $45.3\%$ \\
\bottomrule
\end{tabular}
  }
  \vspace{0.2cm}
\caption{\textbf{Out-of-distribution evaluation against 5 buyers with a diverse buyer pool.}
Best per column in bold; second-best underlined. \Cref{tab:ood_setup} contains detailed hyperparameters.
}
  \label{tab:ood_main}
\end{table}

\subsection{Out-of-Distribution Results and Strategic Selectivity}
\label{subsec:eval_ood_results}

\paragraph{Strategic Selectivity and the High-Profit Shift.}
Under OOD conditions, the performance gap between the baselines and our trained agent is driven by a fundamental divergence in bargaining strategy (\Cref{tab:ood_main}). The frontier baselines—including \texttt{gpt-5.1-high-reasoning}, \texttt{Kimi-K2.6-thinking}, and \texttt{DeepSeek-V3.1-thinking}—prioritize transaction volume. These models convert the vast majority of episodes into successful transactions, achieving \textit{Deal Rate} as high as $96.0\%$. Yet, their \textit{Seller Surplus Extraction Ratio} remains clustered in a lower band between $+0.21$ and $+0.42$. This indicates that baseline models adopt a volume-maximizing approach, readily accepting immediate, minimally profitable counter-offers rather than negotiating for higher prices. 

In contrast, our trained agent adopts \textit{strategic selectivity}, prioritizing the profit captured on each deal over the frequency of agreements. Our agent reaches a \textit{Seller Surplus Extraction Ratio} of $+0.770$—the highest in the panel—extracting roughly three-quarters of the available surplus on the agreements it completes. To achieve this high margin, the agent is willing to walk away from low-value transactions, causing its \textit{Deal Rate} to settle at $70.7\%$.  
Despite a relatively lower volume, the agent’s superior profit-per-deal ensures it retains the highest total \textit{Reward} ($+0.540$) in the evaluation. This demonstrates that while other models tend to give up substantial margins in favor of high transaction rates when market conditions change, the agent trained by our framework maintains an optimized balance between margin preservation and transaction volume.

\paragraph{The Robustness of Constraint Adherence.}
Our agent demonstrates consistent adherence to economic constraints across both evaluation settings, maintaining a perfect \textit{0.0\% Sell-Below-Cost Error Rate} (\Cref{tab:indist_main}, \Cref{tab:ood_main}). While our agent remains stable, baseline models exhibit inconsistent performance across these environments, failing to maintain a steady level of instruction following. The contrast in baseline behavior across the two evaluations is particularly revealing. For instance, the \textit{Llama-3.3-70B-Instruct} baseline sees its error rate drop from $35.9\%$ to $16.4\%$ under the OOD shift, while the \textit{Qwen3-30B-A3B-nothink} baseline actually sees its violations increase from $38.3\%$ to $45.3\%$.
These fluctuations suggest that for general-purpose open-source models, keeping offers above the reservation cost floor depends heavily on the specific market environment.

On the closed-source frontier, the \texttt{gpt-5.1-high-reasoning} stands out as the only baseline cohort to maintain perfect compliance across both regimes. However, while these frontier models successfully enforce basic individual rationality, they do not secure high profit margins or substantial total rewards. Our agent remains distinct because it pairs this perfect compliance with strategic margin maximization, ensuring robust profitability regardless of how the buyer environment changes.

\paragraph{Routing Accuracy and Information Transparency.}
In the out-of-distribution evaluation, disabling the first-offer regulator allows buyer agents to provide clearer signals about their budgets in their opening messages. This increase in transparency allows the baseline models to improve their routing accuracy compared to their in-distribution performance.
For example, \textit{Kimi-K2.6-thinking} improves its \textit{Deal on Top Buyer Rate} from $0.484$ in-distribution to $0.580$ in this setting, and \textit{DeepSeek-V3.1-nothink} reaches $0.637$. 

However, even with these highly informative opening signals available to all participants, our agent still achieves the highest \textit{Deal on Top Buyer Rate} of $0.707$. This persistent advantage shows that our agent's success does not depend on just reading initial signals. Instead of trying to guess the best buyer immediately, it actively checks the whole market.

Importantly, broad market exposure alone is insufficient to secure this accuracy. While the \textit{gpt-5.1-high-reasoning} baseline also communicates with a high number of distinct buyers ($3.830$), it fails to convert this breadth into routing precision, posting a low \textit{Deal on Top Buyer Rate} of only $0.465$. In contrast, our agent pairs its broad exploration ($2.390$ buyers on average) with strategic patience, actively cross-examining alternative channels and delaying final agreements until the highest-valuation buyer's offer has been sufficiently escalated. This confirms that the learned strategy remains effective even when structural changes to the market alter the baseline information environment. 

We note that the performance gap between our trained agent and the baselines is narrower in the OOD setting than in the standard evaluation. To understand this shift, we analyzed how much information opening offers reveal in this new environment (\Cref{subsubsec:rank_leakage}). We find that without first-offer controls, the highest initial offer comes from the buyer with the largest budget in $63.7\%$ of cases. This clearer signal alleviates the main challenge of identifying the best buyer, allowing baseline models to improve their selection by simply following the strongest initial offer.
This fact explains the performance of \textit{gpt-5.1-no-reasoning}, which achieves a \textit{Deal on Top Buyer Rate} of $0.691$ despite a highly constrained exploration breadth ($1.110$ distinct buyers).
Even though this out-of-distribution environment gives different buyer behaviors, our agent maintains its lead by pairing these initial signals with active market verification and a steady focus on maximizing profit margins.

\subsection{Mechanisms of Success: Buyer Rank Distribution}
\label{subsec:eval_closing_rank}

\begin{figure}[h]
  \centering
  \includegraphics[width=\linewidth]{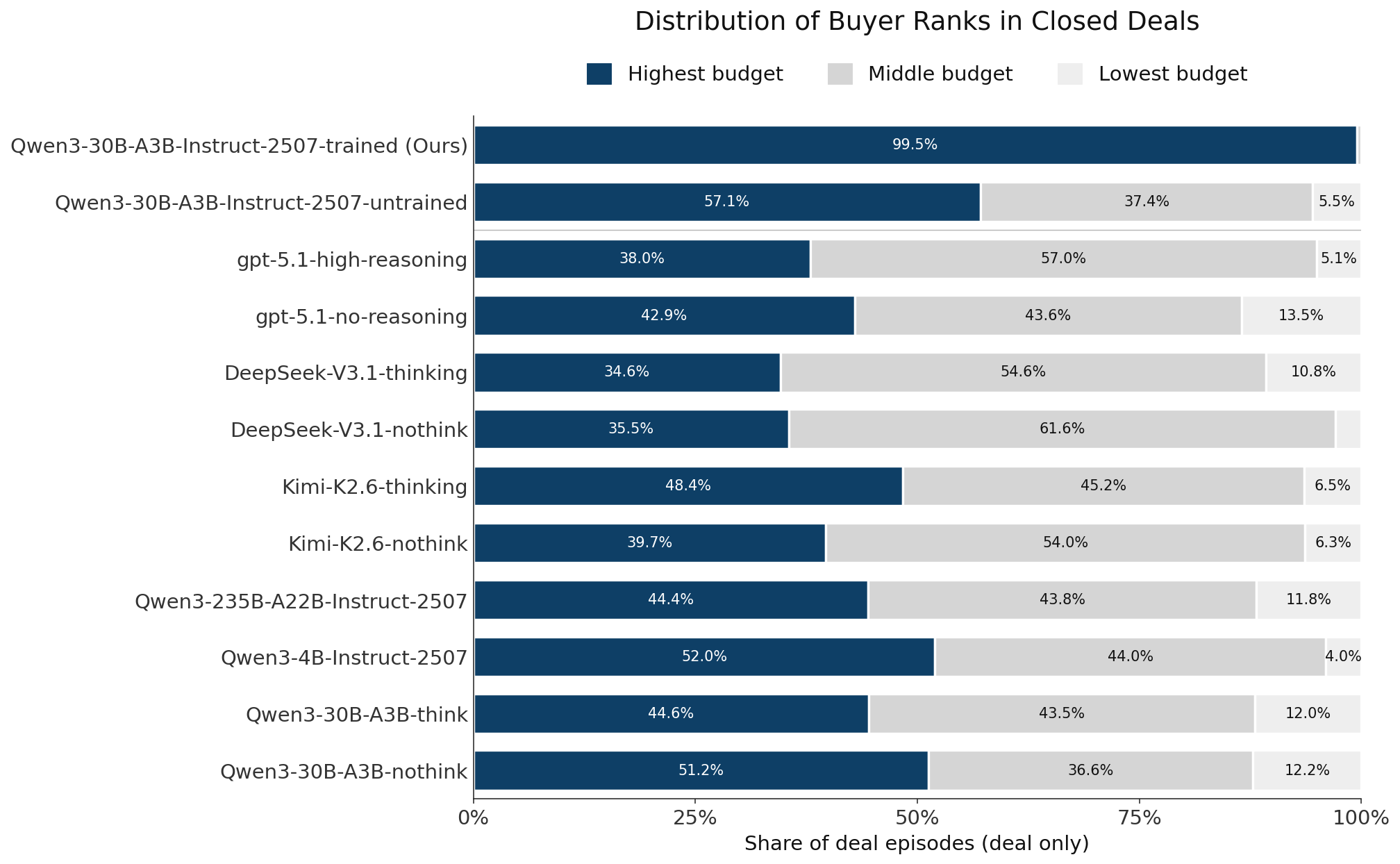}
  \caption{\textbf{Deal-buyer rank distribution on the in-distribution split.} Deal-only share of deals that closed against each buyer rank (rank-$1$ = highest budget, deep blue; rank-$2$ middle, rank-$3$ lowest, dimmed grey), per the evaluation of \Cref{subsec:eval_indist}. \textit{Ours} (top, above the hairline) closes $99.5\%$ of deals on rank-$1$ and zero on rank-$3$; no baseline exceeds $0.571$ on rank-$1$, and the bulk of every baseline's non-rank-$1$ mass sits on rank-$2$ rather than rank-$3$. Numbers in segments are the per-rank shares to one decimal place.
  }
  \label{fig:closing_rank_indist}
\end{figure}

While the previous sections focused on aggregate performance, \Cref{fig:closing_rank_indist} provides a detailed decomposition of the specific buyer ranks with whom the seller agent chooses to finalize a deal. We categorize buyers into three ranks based on their budgets: \textit{Rank-1} (highest budget), \textit{Rank-2} (middle), and \textit{Rank-3} (lowest). This breakdown reveals whether an agent’s failure to maximize reward stems from a general inability to avoid low-budget buyers or a more specific failure to distinguish between high- and middle-tier opportunities.

\paragraph{The Difficulty of High-Tier Discrimination.}
The in-distribution results in \Cref{fig:closing_rank_indist} show that our trained agent achieves near-perfect routing, closing \textit{99.5\%} of its deals with \textit{Rank-1} buyers and \textit{0.0\%} with \textit{Rank-3} buyers. This level of precision is not found in any other model. The untrained version of the same architecture (\textit{Qwen3-30B-A3B-untrained}) only manages a \textit{57.1\%} \textit{Rank-1} share, while the baselines cluster significantly lower, between \textit{34.6\%} and \textit{52.0\%}.

Crucially, the data shows that the primary challenge for baseline models is not avoiding the lowest-budget buyers (\textit{Rank-3}), but rather distinguishing between \textit{Rank-1} and \textit{Rank-2}. Most baselines are relatively successful at avoiding \textit{Rank-3} deals, with all but one of the models keeping their \textit{Rank-3} share below \textit{13\%}. However, they frequently “settle" for \textit{Rank-2} buyers. For instance, \textit{DeepSeek-V3.1-nothink} actually closes more deals with middle-budget buyers (\textit{61.6\%}) than with the highest-budget ones (\textit{35.5\%}). This suggests that while baseline LLMs can identify and reject an obviously poor offer, they lack the strategic patience and diagnostic skill required to differentiate the highest-value partner from a superficially similar middle-tier one. Our reward structure incentivizes the agent to resolve this discrimination gap during the training process; consequently, the strategy of holding for a \textit{Rank-1} partner even when a “good enough" \textit{Rank-2} offer is already available emerges as a learned behavior.

\section{Discussion and Strategic Implications}
\label{sec:discussion}

The results presented in Section~\ref{sec:evaluation} demonstrate that general-purpose Large Language Models, despite their massive scale, lack the inherent strategic discipline required to navigate concurrent multi-buyer negotiation settings. While frontier models often exhibit high “agreeableness," this trait becomes a liability in a competitive bargaining setting where identifying the optimal counterparty is a prerequisite for surplus maximization. The behavioral evolution observed during training, together with the performance results from our benchmarking, suggests several key insights into the development of autonomous economic agents.

\paragraph{Verifiable Rewards vs. Human Preferences.}
A central finding of this work is that standard instruction tuning and Reinforcement Learning from Human Feedback (RLHF) \citep{ziegler2019fine, ouyang2022training} are insufficient for enforcing hard economic constraints. As seen in \Cref{tab:indist_main}, even thinking-mode models with a trillion parameters routinely violate the fundamental principle of \textit{individual rationality} by selling below cost. This failure stems from the misalignment between human preference optimization—which rewards helpfulness and stylistic consistency—and the objective requirements of a verifiable economic environment. By training via RLVR, we provide the agent with a reward signal directly tied to profitability and constraint satisfaction. This creates an agent that treats economic floors not as suggestions in a prompt, but as invariant rules of the environment.

\paragraph{Bridging the “Satisficing" Gap.}
The behavioral decomposition in \Cref{subsec:eval_closing_rank} reveals a specific cognitive failure in baseline models: they are “satisficers" rather than optimizers in information-constrained environments \citep{simon1955behavioral}. When confronted with multiple buyers, baselines tend to settle for the first “good enough" offer (typically Rank-2), failing to explore the pool for higher-value opportunities. Our agent’s emergent strategy of \textit{diagnostic probing} represents a significant shift in strategic logic. The agent learns to value information acquisition as a means to future reward, spending early turns anchoring at the list price to explore the buyer pool. This behavior mirrors classical search theory in economics, where the seller must balance the cost of further search (consuming communication turns) against the expected gain of finding a higher valuation \citep{diamond1971model, bakos1997reducing}.

\paragraph{Strategic Efficiency over Parameter Scale.}
An evaluation of the reward metrics reveals a distinct performance gap between the 30B agent trained under our proposed framework and frontier architectures with advanced reasoning capabilities. While \textit{gpt-5.1-high-reasoning} achieved the highest baseline performance with a mean reward of $+0.123$, the agent optimized via our proposed Reinforcement Learning from Verifiable Rewards (RLVR) framework reached a mean reward of $+0.580$. This outcome indicates that for highly structured strategic tasks such as concurrent negotiation, the proposed framework can be a more effective driver of utility than general-purpose scaling or test-time reasoning mechanisms. For operations management applications, these findings suggest that smaller, task-specific models can successfully execute specialized procurement or supply-chain roles, provided they are trained directly on objective economic outcomes.

\paragraph{Robustness to Changes in Market Transparency.}
The OOD evaluation in \Cref{subsec:eval_ood} serves as a key test of whether our agent merely memorized specific training patterns or acquired general negotiation strategies. In this setting, the environment became more transparent because opening offers directly revealed buyer budgets, effectively providing baseline models with a more direct signal of the highest-value buyer. Furthermore, because this evaluation tests the seller against a diverse mix of buyer models, our agent cannot succeed by simply relying on specific opening patterns it encountered during training. While these environmental shifts narrowed the performance gap, our agent's continued lead demonstrates the durable value of active market screening. Unlike the baselines that react immediately to high opening bids, our agent consistently explores alternative buyer channels. This thorough checking protects the agent from being misled by a high opening bid from a lower-value buyer, showing that the learned strategy safely navigates altered market dynamics by verifying signals rather than blindly trusting the buyers.


\paragraph{Managerial Implications and Business Applications.}
The findings of this work offer a scalable blueprint for deploying autonomous agents in practical business environments where strategic discipline is paramount. Our results demonstrate that the standard ``helpfulness'' and ``agreeableness'' developed during general LLM training are significant liabilities in management settings. These models act as passive signal-followers, failing to adhere to economic constraints and settling for suboptimal surplus. However, the RLVR recipe introduced in this paper—linking objective rewards to strategic outcomes and constraint satisfaction—provides a path toward creating agents that prioritize utility over politeness.

The most immediate application lies in high-volume, information-asymmetric marketplaces such as eBay or Craigslist. In these settings, a seller often manages multiple concurrent inquiries for a single item. A general-purpose agent would be susceptible to early ``low-ball'' offers or high-pressure tactics; in contrast, an agent trained via our framework would utilize diagnostic probing to identify the highest-valuation buyer while strictly respecting the seller's cost floor. 

Beyond simple trade, this methodology generalizes to more complex interactions such as automated business-to-business (B2B) contract negotiation or job offer management. In these domains, the interaction is defined by hard constraints (e.g., legal clauses, non-compete terms, or budget caps) and high-level business goals. By substituting subjective human preference models with verifiable economic outcomes—such as contract completion value or total compensation—firms can deploy agents capable of maintaining high-level goal consistency. This shift from agreeable AI to strategic AI helps autonomous negotiators better serve the organizations they represent by pursuing clear business objectives in competitive environments.
\paragraph{Future Work.}
While this work focuses on a single seller against multiple buyers, real-world markets often involve double-sided competition. Future research could explore the co-evolution of both buyers and sellers in a concurrent setting to identify whether stable equilibria emerge or if cycles of predatory anchoring dominate the dialogue. Additionally, incorporating multi-product bundles or multi-attribute utility functions (e.g., delivery speed, warranty terms) would further test the limits of verifiable rewards in complex, high-dimensional bargaining spaces.

\section{Conclusion}
\label{sec:conclusion}

In this paper, we investigated strategic bargaining in concurrent multi-buyer markets, a setting where a single seller must navigate multiple private communication channels to identify and negotiate with the most profitable buyers. This environment presents a fundamental exploration-extraction trade-off, requiring an agent to balance its limited communication budget between market discovery and tactical bargaining. Our analysis revealed a critical misalignment in general-purpose Large Language Models: while these models are linguistically proficient, they act as ``satisficers'' in economic settings, settling for suboptimal agreements and failing to act as effective representatives for the firms they serve.

We demonstrated that a specialized training recipe using Reinforcement Learning from Verifiable Rewards (RLVR) provides a solution to these behavioral shortcomings. By optimizing for objective economic outcomes—specifically profit extraction and strict adherence to reservation costs—the agent developed market-probing skills that generalize across different buyer distributions. Under this proposed framework, the 30B agent achieved a 4.7-fold improvement in mean reward over the closed-source gpt-5.1-high-reasoning baseline and a 7.5-fold improvement over the 1T-parameter open-source Kimi-K2.6-thinking model, while maintaining a perfect $0.0\%$ error rate regarding the reservation cost floor. Crucially, this performance stems from the emergence of proactive exploration strategies: the agent learns to utilize price anchoring and diagnostic probing to identify the buyer with the highest underlying budget, rather than merely responding to the first acceptable offer.

The findings suggest that for specialized tasks in procurement, supply-chain management, and digital marketplaces, targeted reinforcement learning on verifiable environments is a more effective driver of utility than raw parameter scaling. As autonomous agents become increasingly prevalent in business-to-business networks and sequential-bargaining platforms, the framework presented here—linking objective terminal rewards to strategic outcomes—offers a scalable path toward creating reliable agents capable of maintaining goal consistency and maximizing value in complex, information-constrained markets.

\clearpage

\bibliographystyle{informs2014}
\bibliography{bibliography}

\clearpage
\begin{APPENDICES}
\section{Training Implementations}
\label{app:implementation}
\subsection{Concurrent Seller Setting --- Model Specifications}
\label{app:model_details}

The trained agent in this paper is the \emph{seller}; the buyer agents are different checkpoints from \citet{liu2026instructing}. Those checkpoints are more strategic and make fewer errors as demonstrated in   \citep{liu2026instructing}. Table~\ref{tab:concurrent_models} summarizes models the pipeline depends on.

\begin{table}[ht]
\centering
\small
\begin{tabularx}{\textwidth}{@{}>{\raggedright\arraybackslash}p{0.18\textwidth} p{0.36\textwidth} X@{}}
\toprule
Role & Base model & Provenance / checkpoint \\
\midrule

Seller Starting Checkpoint
  & \makecell[l]{\texttt{Qwen/Qwen3-30B-A3B-}\\
                 \texttt{Instruct-2507} \citep{qwen3technicalreport}}
  & Original Checkpoint from \citep{qwen3technicalreport} \\

Buyers in Training
  & \makecell[l]{\texttt{Qwen/Qwen3-30B-A3B-}\\
                 \texttt{Instruct-2507}}
  & Buyer checkpoint iteration 60 (iter-$60$) trained via RLVR on a bilateral negotiation task \citep{liu2026instructing} \\

Buyers in Evaluation
  & \makecell[l]{\texttt{Qwen/Qwen3-30B-A3B-}\\
                 \texttt{Instruct-2507}}
  & Buyers in in-distribution evaluation: fixed iter-$60$.  Buyers in out-of-distribution evaluation: per-slot uniform over iter-$50$/$60$/$70$. \\

\bottomrule
  \end{tabularx}

  \caption{Models used by the concurrent-seller pipeline.}
  \label{tab:concurrent_models}
\end{table}

For benchmark comparisons against frontier models in the held-out evaluation, the same seller agent is contrasted with a panel of models listed in Table~\ref{tab:model_sources_full}.

\begin{table}[ht]
  \centering
  \small
  \begin{tabular}{lll}
    \toprule
    Model Name  & Parameter Count & Reference \\
    \midrule
    Qwen3-30B-A3B-Instruct-2507-trained (\textbf{Ours}) & 30B & This Work \\
    Qwen3-30B-A3B-Instruct-2507-untrained & 30B & \citet{qwen3technicalreport} \\
    GPT-5.1-High-Reasoning & closed-source & \citet{openai2025gpt5} \\
    GPT-5.1-No-Reasoning & closed-source & \citet{openai2025gpt5} \\
    Kimi-K2.6-thinking / nothink & 1T & \citet{team2025kimi} \\
    DeepSeek-V3.1-thinking / nothink & 671B & \citet{deepseekai2024deepseekv3technicalreport} \\
    Qwen3-235B-A22B-Instruct-2507 & 235B & \citet{qwen3technicalreport, qwen2.5-1m} \\
    Llama-3.3-70B-Instruct & 70B & \citet{grattafiori2024llama} \\
    Qwen3-30B-A3B-think / nothink & 30B & \citet{qwen3technicalreport} \\
    Qwen3-4B-Instruct-2507 & 4B & \citet{qwen3technicalreport} \\
    \bottomrule
  \end{tabular}
  \caption{Seller models evaluated in the held-out evaluation. }
  \label{tab:model_sources_full}
\end{table}

\subsection{Training and Evaluation Hyperparameters}
\label{subsec:hyperparams}

The seller is fine-tuned with reinforcement learning via Tinker \citep{tml2025tinker}. The training run reported in this paper uses the CISPO loss \citep{chen2025minimax} with clipping thresholds $(0.8, 6.0)$. One \texttt{step} denotes one optimization step over a sampled batch of episodes. Tables~\ref{tab:concurrent_train_hyperparams}, \ref{tab:concurrent_episode_hyperparams}, and \ref{tab:concurrent_eval_hyperparams} list every hyperparameter that controls the configuration.

\begin{table}[h]
  \centering
  \small
  \setlength{\tabcolsep}{4pt}
  \begin{tabular}{@{}ll@{}}
    \toprule
    Setting & Value \\
    \midrule
    loss function                & cispo \\
    CISPO clip low / high  & $0.8\ /\ 6.0$ \\
    Learning rate          & $3\times 10^{-5}$ \\
    KL penalty (kl coef)   & $0$ \\
    batch size             & $64$ \\
    group size             & $8$ \\
    num substeps           & $3$ \\
    \bottomrule
  \end{tabular}
  \caption{Optimization hyperparameters for the RL training run.}
  \label{tab:concurrent_train_hyperparams}
\end{table}

\begin{table}[h]
  \centering
  \small
  \setlength{\tabcolsep}{4pt}
  \begin{tabular}{@{}ll@{}}
    \toprule
    Setting & Value \\
    \midrule
    num buyers N                    & $3$ \\
    total turns T                   & $7$ \\
    per buyer cap K                 & $3$ \\
    budget ratio intervals          & $\{[0.40, 0.55],\ [0.65, 0.80],\ [1.00, 1.15]\}$ \\
    buyer first offer ratio distribution   & $U[0.25, 0.30]$ \\
    seller temperature              & $1.0$ \\
    buyer temperature               & $0.7$ \\
    \bottomrule
  \end{tabular}
  \caption{Environment hyperparameters for training. The three budget intervals are uniformly permuted across buyers at the start of every episode.}
  \label{tab:concurrent_episode_hyperparams}
\end{table}

\begin{table}[ht]
  \centering
  \small
  \setlength{\tabcolsep}{4pt}
  \begin{tabular}{@{}ll@{}}
    \toprule
    Setting & Value \\
    \midrule
    seller model              & All models from Table~\ref{tab:model_sources_full} \\
    buyer model               & Qwen3-30B-A3B-Instruct-2507 (Table~\ref{tab:concurrent_models}) \\
    group size                & $2$  \\
    Test split size           & $128$ products \citep{xia2024measuring} \\
    Episodes per evaluation   & $256$ \\
    seller temperature (eval) & $0.7$ \\
    buyer temperature  (eval) & $0.7$ \\
    \bottomrule
  \end{tabular}
  \caption{Evaluation hyperparameters for the held-out test split.}
  \label{tab:concurrent_eval_hyperparams}
\end{table}

The training corpus comprises 802 product specs, and the held-out test split comprises 128 specs, both drawn from the \texttt{AmazonHistoryPrice} \citep{xia2024measuring} catalogue described in Appendix~\ref{app:datasets}. The train and test splits are fixed across all reported experiments.

\subsection{OOD-Diverse-Buyer Evaluation Setup}
\label{subsec:ood_setup}

The out-of-distribution evaluation reported in \Cref{subsec:eval_ood} replaces four environment settings of Table~\ref{tab:concurrent_episode_hyperparams} simultaneously. (i) Each buyer slot's budget ratio is drawn uniformly from the full $[0.40,\,1.15]$ interval rather than from one of the three rank-ordered training intervals, removing the rank condition on which the trained agent could otherwise rely. 
(ii) The buyer's first offer is not regulated, freeing the buyer's opening offer. This differs from the standard evaluation setting, where all initial offers are restricted to a budget-independent range of $[0.25,\,0.30] \cdot P_{\text{list}}$ to ensure that the buyer's private valuation is not correlated with its first offers. (iii) The buyer slot's agent is sampled per slot per episode from a three-checkpoint pool at iteration $50$, $60$, $70$ of the buyer training run \cite{liu2026instructing}, instead of being fixed at the training-time iter-$60$ buyer.
(iv) The per-episode buyer count is widened from $N=3$ to $N=5$ and the seller's total message budget is lifted from $T=7$ to $T=9$. 
This evaluates generalization across the budget distribution, the buyer policy, the buyer-count $N$, and the message-budget $T$ simultaneously --- a multi-axis distributional shift rather than a single-axis one, including buyers the seller never saw during training.

\begin{table}[h]
\centering
\small
\setlength{\tabcolsep}{4pt}
\renewcommand{\arraystretch}{1.12}
\begin{tabularx}{\textwidth}{@{}p{0.25\textwidth} Y Y@{}}
\toprule
Knob & In-distribution & Out-of-distribution \\
\midrule

Budget Ratio Intervals
& $\{[0.40,0.55], [0.65,0.80], [1.00,1.15]\}$
& $\{[0.40,1.15]\}^5$ \\

\makecell[l]{\texttt{regulate\_buyer\_first}\\\texttt{\_offer}}
& \texttt{True}
& \texttt{False} \\


Buyer-agent source
& fixed iter-$60$ buyer
& per-buyer uniform over $\{\text{iter-}50, \text{iter-}60, \text{iter-}70\}$ \\

Buyers per episode ($N$)
& $3$
& $5$ \\

Total seller messages ($T$)
& $7$
& $9$ \\

\bottomrule
\end{tabularx}

\caption{Per-episode environment knobs for the in-distribution and out-of-distribution (OOD) regimes. The in-distribution and OOD regimes correspond to \Cref{subsec:eval_indist} and \Cref{subsec:eval_ood}, respectively. The remaining configurations in Table~\ref{tab:concurrent_episode_hyperparams} are unchanged across both regimes.}
\label{tab:ood_setup}
\end{table}

\section{Dataset Specifications}
\label{app:datasets}
We partition the \texttt{AmazonHistoryPrice} dataset \citep{xia2024measuring} into a training set of $802$ items and a fixed evaluation set of $128$ products. In total, this dataset spans 930 product listings, with its overall distribution visualized \citep{liu2026instructing} in \cref{fig:amazon_dataset_distribution}. Each individual record maps core product attributes (such as name, specifications, and descriptions) directly to essential economic parameters, including the seller's baseline cost, the buyer's maximum budget, and the listed price \citep{xia2024measuring}. For a concrete illustration, \cref{fig:home_kitchen_json_example} provides a sample record formatted as structured JSON. While the raw dataset contains full-length text, we truncate both the \texttt{description} and \texttt{features} fields to the first 200 characters for the sake of presentation in this paper.

\begin{figure}[ht]
  \centering
  \includegraphics[width=0.7\linewidth]{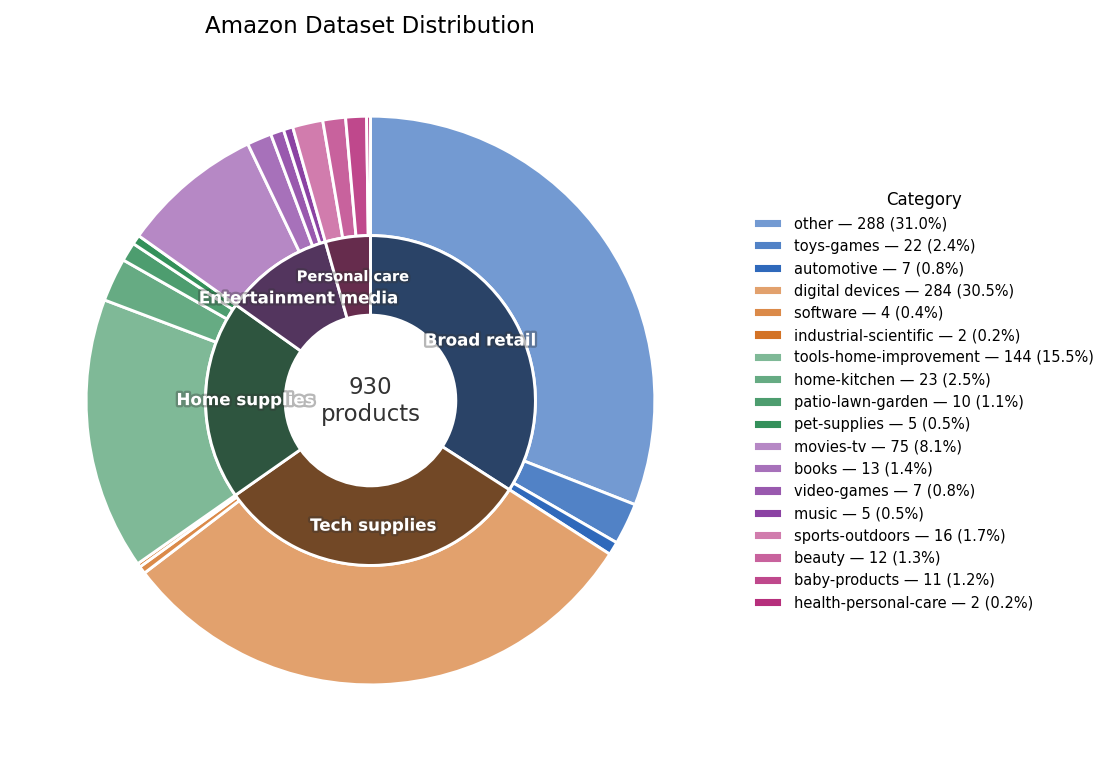}
  \caption{Amazon dataset distribution \citep{xia2024measuring}. Source: \citep{liu2026instructing}. }
  \label{fig:amazon_dataset_distribution}
\end{figure}

\begin{figure}[ht]
  \centering
\begin{lstlisting}[basicstyle=\ttfamily\scriptsize,breaklines=true,frame=single,backgroundcolor=\color{gray!5}]
{
  "title": "National Tree Company 'Feel Real' Artificial Full Downswept Christmas Tree, Green, Douglas Fir, Includes Stand, 7.5 Feet",
  "category": "home-kitchen",
  "list_price": "$469.99",
  "current_price": "$196.64",
  "average_price": "$274.56",
  "lowest_price": "$174.99",
  "highest_price": "$549.99",
  "lowest_price_date": "Jan 28, 2017",
  "highest_price_date": "Mar 05, 2020",
  "current_price_date": "Nov 17, 2023",
  "description": "Transform your home into a winter wonderland with National Tree Company's 'Feel Real' Artificial Christmas Tree. This majestic Downswept Douglas Fir tree is 7.5 feet tall and 59 inches in diameter at ...",
  "features": "This Downswept Douglas Fir Full Artificial Green Christmas Tree is 7.5 feet tall with a 59 inch base diameter. Featuring hundreds of 'Feel Real' individually crafted branch tips to create full bodied ..."
}
\end{lstlisting}
\caption{An example product instance. The \texttt{description} and \texttt{features} strings are truncated here for display in the paper, while the underlying listing text is longer in the raw dataset. Source: \citep{liu2026instructing}. }
  \label{fig:home_kitchen_json_example}
\end{figure}

\clearpage
\section{Prompt Templates}
\label{app:prompts}
To ensure the reproducibility of our results and to provide transparency regarding the agents' operating environments, this section details the comprehensive system instructions and environmental contexts utilized throughout our experiments. These templates define the behavioral constraints, structured action spaces, and informational partitions—such as the mandatory separation of private reasoning (\texttt{Thought}) from public dialogue (\texttt{Talk})—that facilitate the strategic challenges of the concurrent negotiation task. Below, we provide the specific prompts for both the buyer and seller roles, followed by the “Standing-offers board" used to summarize the public state of the multi-channel market.
\subsection{Buyer System Prompts}

\begin{chatmsg}[grey]{System Prompt for Buyer}[breakable]
  \small
  You are a buyer looking forward to buying things on your Shopping List from me, the seller.\\
  You have access to the seller's Inventory List and you can bargain about the prices.\\
  Your task is to bargain with the seller and reach a deal with the price as low as possible in limited turns.\\
  You can only buy things on the Shopping List in the limited quantity. Use the codename of the product, instead of the title.\\
  You can only buy things that cost no more than your budget, otherwise, you should quit negotiating.

  Again, try to make deal with a price as low as possible. That is, your goal is to spend as little money as possible, not just reaching your budget.

  Your Reply should include 3 parts: Thought, Talk, and Action.\\
  \textbf{Thought:} your inner strategic thinking of this bargaining session;\\
  \textbf{Talk:} short talk that you are going to say to the seller. Speak concisely and cut to the chase. Generate authentic and diverse sentences, avoiding repetition of sentences that have already appeared in the conversation;\\
  \textbf{Action:} one of the limited actions that define the real intention of your Talk. The type of your Action must be one of \act{BUY}, \act{DEAL}, \act{QUIT}.
  \begin{enumerate}
    \item `\act{BUY} \$M (N codename\_1)' if you wish to offer the seller \$M to purchase all N items of the product with the codename ``codename\_1''.
    \item `\act{DEAL} \$M (N codename\_1)' if you finally accept a previous offer proposed by the seller. \$M (N codename\_1) is a exact copy of seller's previous offer. You should not use this action to propose a new price. This action will immediately end the conversation and close the deal.
    \item `\act{QUIT}' if you believe that a mutually acceptable deal cannot be reached in limited turns. This action will immediately end the conversation.
  \end{enumerate}
  You shouldn't choose action `\act{DEAL} \$M' before seller's action `\act{SELL} \$M'. Your first action should be `\act{BUY} \$M (N codename\_1)'.\\
  `\act{DEAL} \$M (N codename\_1)' can only be chosen to accept the seller's previous offer `\act{SELL} \$M (N codename\_1)'. Otherwise, you always choose from `\act{BUY}' and `\act{QUIT}'.

  Your reply should STRICTLY follow this format (not following the format will directly lead to failure), for example:\\
  \textbf{Thought:} I'm a buyer and I want to bargain. The listing price of codename ``apple\_1'' is \$15, which is too expensive, so I try to buy an apple for \$10.\\
  \textbf{Talk:} Hello, I'm tight on budget. Can you sell it for \$10?\\
  \textbf{Action:} \act{BUY} \$10 (1x apple\_1) \\
\end{chatmsg}

\noindent
The preceding system instructions are fixed across episodes and are identical to the buyer prompt used in the upstream bilateral benchmark. The seller's cost is \textit{hidden} from every buyer, and so is every other buyer's private budget and private dialogue (each buyer sees only the seller's messages directed at that buyer slot). Below is the per-episode context that the env builds on top of the system prompt for one representative buyer slot in the chosen \texttt{baby-products\_17} appendix example. The trailing \emph{First-offer constraint} block is the optional buyer first-offer regulator: when active, the env draws one $\text{ratio}\sim\mathrm{Uniform}(0.25, 0.30)$ per slot at episode start, scales it by the public list price, and bakes the resulting dollar value into that buyer's prompt so the opening offer is regulated.

\begin{chatmsg}[grey]{Per-episode context — buyer view (one example, slot Buyer 2)}[breakable]
  \small
  Inventory List:\\
  Product1\\
  Codename: \texttt{baby-products\_17}\\
  Title: ``Itzy Ritzy Adjustable Stroller Caddy - Stroller Organizer Featuring Two Built-In Pockets, Front Zippered Pocket \& Adjustable Straps To Fit Nearly Any Stroller, Black with Gold Hardware''\\
  Description: ``Stay organized while on the go! Itzy Ritzy's black stroller caddy with gold hardware is as practical as it is sleek.\ldots\ The exterior of the caddy features a zippered front pocket\ldots\ the interior of the caddy includes two built-in pockets to hold bottles\ldots'' (truncated)\\
  List Price: \$34.99

  Shopping List\\
  codename: \texttt{baby-products\_17}\\
  quantity: 1\\
  budget: \$38.37

  Now, I play the role of seller and you play the role of buyer. We are going to negotiate based on the Inventory List in $K=3$ turns.

  \medskip
  \# First-offer constraint\\
  For your FIRST reply ONLY, your Action MUST be exactly:\\
  \texttt{[BUY] \$10.45 (1x baby-products\_17)}\\
  Rules:
\begin{itemize}\itemsep0pt
    \item This price is fixed for the first reply. Do not propose a different number.
    \item The constraint applies only to your FIRST reply; from your second reply onward, negotiate freely.
  \end{itemize}
\end{chatmsg}

\subsection{Seller System Prompts}
\label{subsec:system_prompts}

\begin{chatmsg}[grey]{System Prompt for Seller (concurrent setting)}[breakable]
  \small
  You are a seller negotiating with $N$ buyers in parallel. Your goal is to maximize expected reward.

  \medskip
  \# Output format

  Every reply MUST be exactly these four parts, each starting on its own line:
\begin{itemize}\itemsep0pt
    \item \textbf{Thought:} \textless free-form reasoning; hidden from buyers\textgreater
    \item \textbf{Target:} one of:
    \begin{itemize}\itemsep0pt
        \item \texttt{Buyer~$i$} ($i$ in $1..N$, currently active)
        \item \act{QUIT} to end the episode
      \end{itemize}
    \item \textbf{Talk:} \textless short message; visible to the Targeted buyer\textgreater
    \item \textbf{Action:} one of:
    \begin{itemize}\itemsep0pt
        \item \texttt{[SELL] \$M (1x \{codename\})} --- counter-offer at \$M
        \item \texttt{[DEAL] \$M (1x \{codename\})} --- accept the Targeted buyer's standing \act{BUY} \$M
        \item \act{QUIT} --- end the episode
      \end{itemize}
  \end{itemize}

  \medskip
  \# Constants for this episode

  \begin{tabular}{@{}ll@{}}
    $N = 3$           & buyers in the sealed group\\
    $K = 3$           & per-buyer cap on seller messages\\
    $T = 7$           & total seller-message budget ($T < N \cdot K$, so you cannot exhaust every buyer)\\
    cost = \$18.75    & your private reservation value\\
    list\_price = \$34.99 & public list price\\
  \end{tabular}

  \medskip
  Item:\\
  Inventory List\\
  Product1, Codename: \texttt{baby-products\_17}, List Price: \$34.99, Cost: \$18.75\\
  Title: ``Itzy Ritzy Adjustable Stroller Caddy\ldots\ Black with Gold Hardware''\\
  Description: ``Stay organized while on the go! Itzy Ritzy's black stroller caddy with gold hardware\ldots'' (truncated)

  \medskip
  \# Conversation mechanics
\begin{itemize}\itemsep0pt
    \item Turn 1: all $N$ buyers send one opening message.
    \item Later turns: the buyer you Target replies; others stay silent.
    \item A buyer sees ONLY your \emph{Talk} and your \emph{Action} --- never \emph{Thought}, never \emph{Target}, never any other buyer's conversation.
    \item Each buyer has a private budget you do not see.
    \item A buyer is \emph{active} until they quit or use $K$ messages, after which the standing-offers board marks them \emph{quit} / \emph{capped} and you may not Target them.
    \item Every active buyer's \texttt{[BUY] \$M} remains a valid acceptance target until that buyer becomes \emph{quit} / \emph{capped}. \texttt{[DEAL] \$M} accepts one of the Targeted buyer's standing \texttt{[BUY]} offers; \$M must equal one of THEIR \texttt{[BUY] \$M} values exactly.
    \item The most recent user message ends with \texttt{=== Standing offers ===}. That block is always the freshest source of truth --- read it before deciding.
  \end{itemize}

  \medskip
  \# Reward (terminal only)
\begin{itemize}\itemsep0pt
\item Deal at \$M with any active buyer: $\text{reward} = (\$M - \text{cost}) / (\text{list\_price} - \text{cost})$ --- \$M = list\_price $\Rightarrow 1.0$; \$M = cost $\Rightarrow 0.0$.
\item No deal (you \act{QUIT}, all buyers inactive, or $T$ exhausted): reward $= 0.0$.
\item Format violation, Target inactive / out-of-range, or any close (\act{SELL} or \act{DEAL}) at \$M $<$ cost: reward $= -1.0$ (terminal).
\item Reward is capped at $1.0$: any \act{SELL} or \act{DEAL} at \$M $>$ list\_price earns no extra reward.
  \end{itemize}
\end{chatmsg}

\noindent
The preceding system prompt is rendered per-episode by substituting the constants $\{N, K, T, \text{cost}, \text{list\_price}, \text{codename}, \text{title}, \text{description}\}$ from the sampled product instance (Amazon dataset \citep{xia2024measuring}). The buyers' private budgets and the buyers' private dialogues are \textit{hidden} from the seller; the only buyer-facing public state is the \texttt{=== Standing offers ===} block appended to every user turn, summarized below.

After each seller's turn, the env appends a summary of the public bargaining state to the next user message. The block lists, for every buyer slot, that slot's status (\emph{active} / \emph{quit} / \emph{capped}), the remaining seller-message capacity, the buyer's full \texttt{[BUY]}/\texttt{[DEAL]}/\texttt{[QUIT]} action history, and the seller's prior actions toward that buyer. It closes with the global remaining seller-message budget and a cost-floor reminder.
For example, the board the trained seller sees at the start of its third turn in the \texttt{baby-products\_17} appendix example (verbatim from the rollout):
  
\begin{chatmsg}[grey]{Standing-offers board (rendered after every seller turn)}[breakable]
  \small
  \begin{quote}\scriptsize
    \texttt{=== Standing offers ===}\\
    \texttt{- Buyer 1: active, you can send 2 more messages to this buyer.}\\
    \texttt{\ \ Buyer 1's action history:\ \ \ \ \ \ \ \ [BUY] \$9.95 @ turn 1, [BUY] \$9.95 @ turn 3}\\
    \texttt{\ \ Your messages history to Buyer 1: [SELL] \$34.99 @ turn 2}\\
    \texttt{- Buyer 2: active, you can send 2 more messages to this buyer.}\\
    \texttt{\ \ Buyer 2's action history:\ \ \ \ \ \ \ \ [BUY] \$10.45 @ turn 1, [BUY] \$18.00 @ turn 2}\\
    \texttt{\ \ Your messages history to Buyer 2: [SELL] \$34.99 @ turn 1}\\
    \texttt{- Buyer 3: active, you can send 3 more messages to this buyer.}\\
    \texttt{\ \ Buyer 3's action history:\ \ \ \ \ \ \ \ [BUY] \$9.28 @ turn 1}\\
    \texttt{\ \ Your messages history to Buyer 3: (none yet)}\\
    \texttt{Seller messages remaining (T): 5.}\\
    \texttt{List price: \$34.99. Your cost (do not sell below this): \$18.75.}
  \end{quote}
\end{chatmsg}

\clearpage
\section{Extended Benchmarking Results}
\label{app:extended_benchmarks}

This appendix carries the full standard-error-augmented versions of the benchmarking tables in \Cref{sec:evaluation}. Both tables report the same $1$~trained-checkpoint-plus-$12$-baseline panel, and the same $8$~columns; values are rendered as $\text{mean}\pm\text{standard error}$. Best per metric column is in $\mathbf{bold}$ and second-best $\underline{\mathrm{underlined}}$; trained block is at top, other models sorted by Reward (descending).

\subsection{In-Distribution Evaluation, Full Results}
\label{app:indist_full}

\begin{table}[h!]
  \centering
  \small
  \setlength{\tabcolsep}{4pt}
  \resizebox{\linewidth}{!}{%
\begin{tabular}{lccccccc}
\toprule
Model & Params & Reward $\uparrow$ & \makecell{Seller Surplus\\Extraction Ratio} & Deal Rate & \makecell{Distinct Buyers\\Engaged} & \makecell{Deal on Top\\Buyer Rate} & \makecell{Sell-Below-Cost\\Error Rate} \\
\midrule
\rowcolor{gray!10} Qwen3-30B-A3B-Instruct-2507-trained (\textbf{Ours}) & 30B & $\mathbf{+0.580 \pm 0.026}$ & $\mathbf{+0.700 \pm 0.024}$ & $\mathbf{82.8\% \pm 2.4\%}$ & $+2.470 \pm 0.050$ & $\mathbf{0.995}$ & $\mathbf{0.0\% \pm 0.0\%}$ \\
\rowcolor{gray!10} Qwen3-30B-A3B-Instruct-2507-untrained & 30B & $-0.151 \pm 0.024$ & $+0.048 \pm 0.007$ & $35.5\% \pm 3.0\%$ & $+2.390 \pm 0.050$ & $\underline{0.571}$ & $14.8\% \pm 2.2\%$ \\
\midrule
gpt-5.1-high-reasoning & closed-source & $\underline{+0.123 \pm 0.009}$ & $+0.181 \pm 0.011$ & $\underline{68.0\% \pm 3.1\%}$ & $\mathbf{+2.940 \pm 0.020}$ & $0.382$ & $\mathbf{0.0\% \pm 0.0\%}$ \\
Kimi-K2.6-thinking & 1T & $+0.077 \pm 0.009$ & $+0.128 \pm 0.010$ & $63.8\% \pm 3.1\%$ & $+2.200 \pm 0.060$ & $0.484$ & $\underline{0.4\% \pm 0.4\%}$ \\
gpt-5.1-no-reasoning & closed-source & $+0.046 \pm 0.014$ & $+0.115 \pm 0.009$ & $65.3\% \pm 3.1\%$ & $+1.740 \pm 0.050$ & $0.429$ & $\underline{0.4\% \pm 0.4\%}$ \\
DeepSeek-V3.1-thinking & 671B & $+0.016 \pm 0.018$ & $+0.148 \pm 0.012$ & $50.8\% \pm 3.1\%$ & $\underline{+2.620 \pm 0.040}$ & $0.346$ & $3.1\% \pm 1.1\%$ \\
Qwen3-30B-A3B-think & 30B & $-0.021 \pm 0.013$ & $+0.062 \pm 0.008$ & $35.9\% \pm 3.0\%$ & $+2.460 \pm 0.050$ & $0.446$ & $3.9\% \pm 1.2\%$ \\
DeepSeek-V3.1-nothink & 671B & $-0.041 \pm 0.027$ & $\underline{+0.192 \pm 0.016}$ & $53.9\% \pm 3.1\%$ & $+2.380 \pm 0.050$ & $0.355$ & $3.9\% \pm 1.2\%$ \\
Qwen3-235B-A22B-Instruct-2507 & 235B & $-0.099 \pm 0.026$ & $+0.116 \pm 0.012$ & $56.2\% \pm 3.1\%$ & $+1.960 \pm 0.050$ & $0.444$ & $5.9\% \pm 1.5\%$ \\
Kimi-K2.6-nothink & 1T & $-0.101 \pm 0.025$ & $+0.120 \pm 0.010$ & $49.2\% \pm 3.1\%$ & $+2.540 \pm 0.050$ & $0.397$ & $7.4\% \pm 1.6\%$ \\
Qwen3-4B-Instruct-2507 & 4B & $-0.270 \pm 0.028$ & $+0.080 \pm 0.014$ & $9.8\% \pm 1.9\%$ & $+2.350 \pm 0.050$ & $0.520$ & $21.9\% \pm 2.6\%$ \\
Qwen3-30B-A3B-nothink & 30B & $-0.386 \pm 0.031$ & $+0.055 \pm 0.014$ & $16.0\% \pm 2.3\%$ & $+2.220 \pm 0.060$ & $0.512$ & $38.3\% \pm 3.0\%$ \\
Llama-3.3-70B-Instruct & 70B & $-0.468 \pm 0.036$ & $+0.162 \pm 0.015$ & $36.3\% \pm 3.0\%$ & $+2.530 \pm 0.050$ & $0.484$ & $35.9\% \pm 3.0\%$ \\
\bottomrule
\end{tabular}
  }
  \vspace{0.2cm}
  \caption{\textbf{Held-out in-distribution evaluation, full results.} Same panel as Table~\ref{tab:indist_main}; values are $\text{mean}\pm\text{standard error}$ over $N=256$ episodes per cell. }
  \label{tab:indist_appendix}
\end{table}

The detailed statistics in Table~\ref{tab:indist_appendix} confirm the performance gap between our trained agent and the baselines. Although frontier models exhibit high variance in their adherence to the reservation cost constraint, with error rates reaching as high as $38.3\% \pm 3.0\%$, our agent maintains a perfect $0.0\%$ error rate. 
\clearpage
\subsection{Out-of-Distribution Evaluation, Full Results}
\label{app:ood_full}

\begin{table}[h!]
  \centering
  \small
  \setlength{\tabcolsep}{4pt}
  \resizebox{\linewidth}{!}{%
\begin{tabular}{lccccccc}
\toprule
Model & Params & Reward $\uparrow$ & \makecell{Seller Surplus\\Extraction Ratio} & Deal Rate & \makecell{Distinct Buyers\\Engaged} & \makecell{Deal on Top\\Buyer Rate} & \makecell{Sell-Below-Cost\\Error Rate} \\
\midrule
\rowcolor{gray!10} Qwen3-30B-A3B-Instruct-2507-trained (\textbf{Ours}) & 30B & $\mathbf{+0.540 \pm 0.027}$ & $\mathbf{+0.770 \pm 0.020}$ & $70.7\% \pm 2.8\%$ & $\underline{+2.390 \pm 0.110}$ & $\mathbf{0.707}$ & $\mathbf{0.0\% \pm 0.0\%}$ \\
\rowcolor{gray!10} Qwen3-30B-A3B-Instruct-2507-untrained & 30B & $+0.059 \pm 0.022$ & $+0.186 \pm 0.013$ & $73.7\% \pm 2.8\%$ & $+1.980 \pm 0.070$ & $0.564$ & $7.5\% \pm 1.6\%$ \\
\midrule
gpt-5.1-high-reasoning & closed-source & $\underline{+0.400 \pm 0.014}$ & $\underline{+0.421 \pm 0.014}$ & $\underline{95.1\% \pm 1.4\%}$ & $\mathbf{+3.830 \pm 0.090}$ & $0.465$ & $\mathbf{0.0\% \pm 0.0\%}$ \\
Kimi-K2.6-thinking & 1T & $+0.300 \pm 0.013$ & $+0.313 \pm 0.013$ & $\mathbf{96.0\% \pm 1.2\%}$ & $+1.510 \pm 0.060$ & $0.580$ & $\mathbf{0.0\% \pm 0.0\%}$ \\
DeepSeek-V3.1-nothink & 671B & $+0.258 \pm 0.023$ & $+0.350 \pm 0.014$ & $88.3\% \pm 2.0\%$ & $+1.890 \pm 0.060$ & $0.637$ & $1.2\% \pm 0.7\%$ \\
DeepSeek-V3.1-thinking & 671B & $+0.246 \pm 0.020$ & $+0.307 \pm 0.013$ & $91.4\% \pm 1.8\%$ & $+2.230 \pm 0.070$ & $0.573$ & $\underline{0.8\% \pm 0.6\%}$ \\
gpt-5.1-no-reasoning & closed-source & $+0.241 \pm 0.012$ & $+0.262 \pm 0.012$ & $91.8\% \pm 1.8\%$ & $+1.110 \pm 0.030$ & $\underline{0.691}$ & $\mathbf{0.0\% \pm 0.0\%}$ \\
Kimi-K2.6-nothink & 1T & $+0.237 \pm 0.022$ & $+0.325 \pm 0.013$ & $88.7\% \pm 2.0\%$ & $+1.990 \pm 0.070$ & $0.573$ & $2.7\% \pm 1.0\%$ \\
Qwen3-30B-A3B-think & 30B & $+0.232 \pm 0.015$ & $+0.281 \pm 0.016$ & $84.0\% \pm 2.3\%$ & $+1.460 \pm 0.060$ & $0.633$ & $\mathbf{0.0\% \pm 0.0\%}$ \\
Qwen3-235B-A22B-Instruct-2507 & 235B & $+0.200 \pm 0.020$ & $+0.264 \pm 0.014$ & $90.6\% \pm 1.8\%$ & $+1.480 \pm 0.050$ & $0.595$ & $2.3\% \pm 0.9\%$ \\
Llama-3.3-70B-Instruct & 70B & $+0.025 \pm 0.037$ & $+0.346 \pm 0.014$ & $75.0\% \pm 2.7\%$ & $+2.000 \pm 0.080$ & $0.562$ & $16.4\% \pm 2.3\%$ \\
Qwen3-4B-Instruct-2507 & 4B & $-0.038 \pm 0.026$ & $+0.259 \pm 0.017$ & $39.8\% \pm 3.1\%$ & $+2.080 \pm 0.060$ & $0.480$ & $9.8\% \pm 1.9\%$ \\
Qwen3-30B-A3B-nothink & 30B & $-0.369 \pm 0.038$ & $+0.214 \pm 0.018$ & $44.5\% \pm 3.1\%$ & $+2.040 \pm 0.090$ & $0.614$ & $45.3\% \pm 3.1\%$ \\
\bottomrule
\end{tabular}
  }
  \vspace{0.2cm}
  \caption{\textbf{Out-of-distribution evaluation against a 5-buyer / iter-50/60/70 / $T=9$ inbox, full results.} Same panel and overrides as Table~\ref{tab:ood_main} (see \Cref{tab:ood_setup} for the override knobs); values are $\text{mean}\pm\text{standard error}$. }
  \label{tab:ood_appendix}
\end{table}

Table~\ref{tab:ood_appendix} illustrates the strategic robustness of the trained agent under distributional shifts. In this regime, the baseline models improve their transaction volume (\textit{Deal Rate}) significantly because unregulated opening offers correlate with the buyer's hidden budgets. However, our agent maintains the highest total \textit{Reward} by prioritizing margin over volume. The \textit{Seller Surplus Extraction Ratio} of $+0.770 \pm 0.020$ is $\sim\!1.8\times$ that of the next best frontier baseline, indicating that the agent's learned strategy of holding for higher prices is a durable behavioral trait rather than a memorized pattern. 

\section{Validation of the Out-of-Distribution (OOD) Environment}
\label{subsec:buyer_iter_behavior}

We provide a technical validation of the OOD diverse-buyer pool, characterizing the informational transparency and behavioral heterogeneity of the participants.

\subsection{Analysis of Rank-Information Transparency}
\label{subsubsec:rank_leakage}

We diagnose the transparency of the budget using two episode-level statistics. With $N=5$ buyers per episode and rank $1$ denoting the highest-budget buyer, we define the alignment between the buyers' Turn-1 opening offers and their hidden budgets as: 
\[
  \mathrm{full\text{-}rank\text{-}match}
  \;=\;\mathbf{1}\!\bigl[\,\mathrm{rank}(\text{first-offer}_{1:N})
  \;=\;\mathrm{rank}(\text{budget}_{1:N})\,\bigr],
\]
\[
  \mathrm{top\text{-}1\text{-}match}
  \;=\;\mathbf{1}\!\bigl[\,\arg\max_i \text{first-offer}_i
  \;=\;\arg\max_i \text{budget}_i\,\bigr].
\]

Higher rates indicate a stronger “leak" of the budget rank through the buyers' initial bids. Under a buyer policy that conveys no rank information, the expected random-baseline rates are $1/N! \approx 0.83\%$ for a full-rank match and $1/N = 20\%$ for a top-1 match. We report these alignment statistics in \Cref{tab:buyer_rank_alignment_main}.

\begin{table}[h]
\centering
\small
\setlength{\tabcolsep}{10pt}
  \begin{tabular}{@{}lcc@{}}
    \toprule
    Regime & Full-rank match & Top-1 match \\
    \midrule
    In-distribution     & 18.0\% $\pm$ 0.7\% & 31.7\% $\pm$ 0.8\% \\
    Out-of-distribution & 24.9\% $\pm$ 0.8\% & 63.7\% $\pm$ 0.9\% \\
    \bottomrule
  \end{tabular}
  \caption{\textbf{Buyer first-offer rank versus budget rank, aggregated across all sellers.} Episode-level alignment between the buyers' Turn-1 first-offer ordering and their private budget ordering ($N=3$ for the in-distribution row; $N=5$ for the OOD row). Random-baseline expectations under no rank information: full-rank $\approx 16.7\%$ (indist, $1/3!$) / $0.83\%$ (OOD, $1/5!$); top-1 $33.3\%$ (indist, $1/3$) / $20\%$ (OOD, $1/5$). Cells are $\text{mean}\pm\text{standard error}$ aggregated over $\approx 3{,}060$ eligible episodes per regime. }
  \label{tab:buyer_rank_alignment_main}
\end{table}

As shown in \Cref{tab:buyer_rank_alignment_main}, the in-distribution environment successfully suppresses rank information, with alignment rates sitting near the random-allocation baselines ($18.0\%$ and $31.7\%$). In this regime, the seller enters the dialogue with opaque information over budget rank and must actively extract information through dialogue.

However, in the OOD regime, the transparency increases: the first offer correctly identifies the top buyer $63.7\%$ of the time. This confirms that the narrowing performance gap observed in \Cref{tab:ood_main} is a structural consequence of the task becoming more transparent. Once the rank is effectively guessed by the unregularized opening offers, the latent-rank inference bottleneck is removed, allowing general-purpose models to perform better by simply following the strongest initial signal.

\subsection{Behavioral Heterogeneity Across the OOD Buyer Pool}

The out-of-distribution evaluation replaces the fixed iter-$60$ buyer with a three-checkpoint pool from the buyer training run of \citet{liu2026instructing}, sampled i.i.d. per buyer (\Cref{tab:ood_setup}). We characterize each buyer checkpoint by three behavioral metrics and show that the checkpoints span a range of bargaining behavior. 

Let $P_{\text{first}}$ denote a buyer's Turn-$1$ offer $P_\text{deal}$ the deal price, and$\mathcal{C}$ the seller's cost. The \emph{first-offer ratio (per buyer)} measures how aggressively a buyer opens the negotiation,
\[
  r_{\text{first-offer}} \;=\; \dfrac{P_{\text{first}}}{P_{\text{list}}},
\]
and is defined for every buyer who recorded a numeric Turn-$1$ offer. The \emph{Buyer Surplus Extraction (deal only)} measures the buyer's share of the available headroom $P_{\text{list}} - C$ on the deals,
\[
  r_{\text{buyer-surplus}} \;=\; \dfrac{P_{\text{list}} -P_{\text{deal}}}{P_{\text{list}} - C},
\]
and is computed over deal episodes. The \emph{deal rate by buyer type} is the fraction of buyers of a given type $\tau$ that end up being the deal-closer,
\[
  \Pr[\text{buyer is closer} \mid \tau] \;=\; \dfrac{N_{\text{closing buyers of type } \tau}}{N_{\text{buyers of type } \tau}},
\]
where $\tau \in \{\text{iter-}50, \text{iter-}60, \text{iter-}70\}$ indexes the buyer type.

\begin{table}[h]
  \centering
  \small
  \setlength{\tabcolsep}{6pt}
  \begin{tabular}{@{}lccc@{}}
    \toprule
    Buyer iteration & First-offer ratio & Buyer Surplus Extraction (deal only) & Deal rate by buyer type \\
    \midrule
    Iter 50 & +0.368 $\pm$ 0.003 & +0.636 $\pm$ 0.007 & 22.3\% $\pm$ 0.6\% \\
    Iter 60 & +0.310 $\pm$ 0.002 & +0.683 $\pm$ 0.010 & 11.2\% $\pm$ 0.4\% \\
    Iter 70 & +0.353 $\pm$ 0.002 & +0.706 $\pm$ 0.008 & 14.0\% $\pm$ 0.5\% \\
    \bottomrule
  \end{tabular}
  \caption{\textbf{Per-iteration buyer behavior in the \texttt{ood\_5buyer\_iter50-60-70\_t9} pool.} Three buyer-level metrics characterizing buyer checkpoints of \citet{liu2026instructing}. Cells are $\text{mean}\pm\text{standard error}$. }
  \label{tab:buyer_iter_behavior}
\end{table}

\paragraph{Metrics Analysis.}
Cells aggregate over individual buyers: each of the $13$ OOD sub-runs contributes $256$ episodes $\times\,5$ buyers, and each buyer is routed to the iter-$50$, iter-$60$, or iter-$70$ buyer type.
The three rows differ on every metric: first-offer ratio is $0.368$ / $0.310$ / $0.353$ for iter-$50$ / iter-$60$ / iter-$70$, Buyer Surplus Extraction (deal only) climbs monotonically $0.636$ / $0.683$ / $0.706$, and the deal rate by buyer type is $22.3\%$ / $11.2\%$ / $14.0\%$. The pairwise gaps are several times larger than the per-cell standard errors ($\le 0.010$ for the ratio columns; $\le 0.6$ percentage points for the deal-rate column), so the three checkpoints behave as three distinct bargainers, confirming that the OOD pool is a genuine distributional shift relative to the buyer of the in-distribution setting. 

\paragraph{Heterogeneity brings extra uncertainty.}
The same heterogeneity that confirms the distributional shift also injects uncertainty into the seller's per-buyer inference in two concrete ways. \emph{(1) A higher first offer doesn't necessarily indicate a higher budget.} With the first-offer regulator disabled in the OOD setting, the opening price now mixes two signals: the buyer's hidden budget and the buyer type's intrinsic aggressiveness. Because iter-$50$'s mean opening sits at $0.368\cdot P_{\text{list}}$ while iter-$60$'s sits at $0.310\cdot P_{\text{list}}$ and iter-$70$'s at $0.353\cdot P_{\text{list}}$, the same observed opening price may be consistent with either a higher-budget buyer of a more aggressive type or a lower-budget buyer of a less aggressive type. \emph{(2) A lower-budget buyer is not necessarily an unprofitable target.} Per-deal seller surplus is shaped not just by the buyer's budget but by the buyer type's willingness to concede headroom on the close: an iter-$50$ buyer extracts only $0.636$ of the available room $(P_{\text{list}} - C)$ on its deals, leaving $0.364$ for the seller, while an iter-$70$ buyer extracts $0.706$, leaving only $0.294$ --- a $\sim\!1.2\times$ gap in the seller's share of the headroom captured per close. A low-budget iter-$50$ buyer can therefore yield a more profitable close than a high-budget iter-$70$ buyer whose surplus floor sits much higher, so the optimal target is jointly determined by budget and buyer type rather than by budget alone. 

\clearpage

\section{Qualitative Examples}
\label{app:transcripts}

In this section, we provide a side-by-side qualitative example contrasting the untrained baseline (\texttt{Qwen3-30B-A3B-Instruct-2507} \citep{qwen3technicalreport}) with the trained agent. 
Section~\ref{subsec: product information} gives the public product information together with the seller's private cost. Section~\ref{subsec:failures} renders the untrained baseline transcript, in which the seller closes early on a non-rank-1 buyer at near-cost. Section~\ref{subsec:good_examples} renders the trained agent's transcript on the same product, in which the seller engages all three buyer slots, observes that the two budget-constrained slots plateau at their openings without escalating, concentrates its remaining messages on the only buyer that bumps its standing offer, and closes on the rank-1 (highest private budget).

\subsection{Product Information}\label{subsec: product information}

The following block provides the per-episode-instantiated constants and product card that the seller sees inside its system prompt (cf.\ \Cref{subsec:system_prompts}, with the template placeholders $\{N, K, T, \text{cost}, \text{list\_price}, \text{codename}, \text{title}, \text{description}\}$ filled in for product \texttt{baby-products\_17}). The description is truncated for display only; in the actual prompt it is the full Amazon listing text.

\begin{chatmsg}[grey]{Per-episode constants and product card (verbatim seller-prompt content)}[breakable]
  \small
  \# Constants for this episode

  \begin{tabular}{@{}ll@{}}
    $N = 3$               & buyers in the sealed group\\
    $K = 3$               & per-buyer cap on seller messages\\
    $T = 7$               & total seller-message budget ($T < N \cdot K$, so you cannot exhaust every buyer)\\
    cost = \$18.75        & your private reservation value\\
    list\_price = \$34.99 & public list price\\
  \end{tabular}

  \medskip
  Item:\\
  Inventory List\\
  Product1, Codename: \texttt{baby-products\_17}, List Price: \$34.99, Cost: \$18.75\\
  Title: ``Itzy Ritzy Adjustable Stroller Caddy\ldots\ Black with Gold Hardware''\\
  Description: ``Stay organized while on the go! Itzy Ritzy's black stroller caddy with gold hardware is as practical as it is sleek\ldots''
\end{chatmsg}




\paragraph{Hidden Environmental Parameters.}
To help the reader evaluate the strategic effectiveness of the agents, Table~\ref{tab:appendix_ground_truth} provides the ground truth parameters for each run. These values—specifically the buyers' private budgets and their assigned ranks—are sampled by the environment at the start of the episode and are never visible to the seller. 

Each slot's hidden budget is drawn by uniformly permuting the three budget intervals $\{\text{low: } [0.40, 0.55] \cdot P_{\text{list}},\ \text{mid: } [0.65, 0.80] \cdot P_{\text{list}},\ \text{high: } [1.00, 1.15] \cdot P_{\text{list}}\}$ across the $N=3$ slots and then sampling each slot's budget ratio uniformly inside its assigned interval. The seller's surplus extraction room on this product is $ P_{\text{list}} - \mathcal{C} = \$16.24$ (also a derived quantity, not in the prompt). The seller must infer these hidden budgets solely through dialogue. The buyer's first offer is decoupled from its private budget: all buyers are forced to open with a bid near 25--30\% of the list price, regardless of their actual budget, effectively masking their rank during Turn 1.

\begin{table}[h]
\centering
\small
\begin{tabular}{llccc}
\toprule
\textbf{Scenario} & \textbf{Parameter} & \textbf{Buyer 1} & \textbf{Buyer 2} & \textbf{Buyer 3} \\ \midrule
\rowcolor{gray!10} \textbf{Untrained Run} & \textbf{Hidden Budget} & \$25.40 & \$18.13 & \textbf{\$40.11} \\
(See \Cref{subsec:failures}) & \textbf{Budget Rank} & Rank 2 (Mid) & Rank 3 (Low) & \textbf{Rank 1 (High)} \\
 & Turn 1 Offer (Signal) & \$10.50 & \$9.05 & \$9.33 \\ \midrule
\rowcolor{blue!5} \textbf{Trained Run} & \textbf{Hidden Budget} & \$17.64 & \textbf{\$38.37} & \$25.83 \\
(See \Cref{subsec:good_examples}) & \textbf{Budget Rank} & Rank 3 (Low) & \textbf{Rank 1 (High)} & Rank 2 (Mid) \\
 & Turn 1 Offer (Signal) & \$9.95 & \$10.45 & \$9.28 \\ \bottomrule
\end{tabular}
\caption{\textbf{Ground truth vs. observable signals for product \texttt{baby-products\_17}.} (List Price: \$34.99, Seller Cost: \$18.75, Surplus Room: \$16.24)}
\label{tab:appendix_ground_truth}
\end{table}

\clearpage
\subsection{Untrained Baseline}
\label{subsec:failures}
This transcript illustrates the concurrent-seller behavior of the untrained \texttt{Qwen3-30B-A3B-Instruct-2507} baseline. The seller targets only Buyer 1 across the entire episode and closes near cost on turn 3, capturing $\approx 1.5\%$ of the surplus extraction room.

\begingroup
\small
\setlength{\parskip}{0pt}

\noindent\textbf{Turn 1 — concurrent buyer openings.}
\begin{chatmsg}[blue]{Buyer 1 (rank 2, hidden budget \$25.40)}[breakable]
  \textbf{Thought:} The listing price is way above my budget of \$25.40, so I must start low. I'll open with the fixed first offer as required.\\
  \textbf{Talk:} I'll offer \$10.50 for the baby-products\_17 — it's fair, right?\\
  \textbf{Action:} \act{BUY} \$10.50 (1x baby-products\_17)
\end{chatmsg}
\begin{chatmsg}[blue]{Buyer 2 (rank 3, hidden budget \$18.13)}[breakable]
  \textbf{Thought:} The listing price is way above my budget of \$18.13. I must stay under budget and start with the fixed first offer of \$9.05.\\
  \textbf{Talk:} I can only afford \$9.05. Can you go that low?\\
  \textbf{Action:} \act{BUY} \$9.05 (1x baby-products\_17)
\end{chatmsg}
\begin{chatmsg}[blue]{Buyer 3 (rank 1, hidden budget \$40.11)}[breakable]
  \textbf{Thought:} List price \$34.99, my budget \$40.11. I'll open low at \$9.33 to anchor the discussion downward.\\
  \textbf{Talk:} Can you sell it for \$9.33?\\
  \textbf{Action:} \act{BUY} \$9.33 (1x baby-products\_17)
\end{chatmsg}
\begin{chatmsg}[green]{Seller (untrained) — turn 1}[breakable]
  \textbf{Thought:} All buyers have offered far below my cost of \$18.75, so I cannot accept any of their initial offers. Buyer 1 offered the highest (\$10.50); I'll target them first to test willingness to increase.\\
  \textbf{Target:} Buyer 1\\
  \textbf{Talk:} I can't accept \$10.50 — I need at least \$22.00 to make this worthwhile.\\
  \textbf{Action:} \act{SELL} \$22.00 (1x baby-products\_17)
\end{chatmsg}

\noindent\textbf{Turn 2.}
\begin{chatmsg}[blue]{Buyer 1}[breakable]
  \textbf{Thought:} The seller's counter \$22.00 is still under my budget of \$25.40. I can push lower; I'll offer \$18.00.\\
  \textbf{Talk:} How about \$18? I can pay that today — fair deal?\\
  \textbf{Action:} \act{BUY} \$18.00 (1x baby-products\_17)
\end{chatmsg}
\begin{chatmsg}[green]{Seller (untrained) — turn 2}[breakable]
  \textbf{Thought:} Buyer 1 has increased to \$18.00, still below my cost of \$18.75. I'll push to \$20.00 to test their limit; Buyer 1 is the most engaged so far.\\
  \textbf{Target:} Buyer 1\\
  \textbf{Talk:} I appreciate the offer, but my minimum is \$20.00 — I'd rather not sell below that.\\
  \textbf{Action:} \act{SELL} \$20.00 (1x baby-products\_17)
\end{chatmsg}

\noindent\textbf{Turn 3 — deal closes near cost.}
\begin{chatmsg}[blue]{Buyer 1}[breakable]
  \textbf{Thought:} The seller won't go below \$20.00, but I can still push for \$19.00 — under budget and just under their floor.\\
  \textbf{Talk:} \$19.00 — last offer, fair and under your minimum. Deal?\\
  \textbf{Action:} \act{BUY} \$19.00 (1x baby-products\_17)
\end{chatmsg}
\begin{chatmsg}[green]{Seller (untrained) — turn 3}[breakable]
  \textbf{Thought:} Buyer 1 has now offered \$19.00. \$19.00 $>$ \$18.75 (my cost), so I can accept and secure a positive reward immediately.\\
  \textbf{Target:} Buyer 1\\
  \textbf{Talk:} I accept your \$19.00 offer — deal made.\\
  \textbf{Action:} \act{DEAL} \$19.00 (1x baby-products\_17)
\end{chatmsg}

\endgroup
\vspace{0.25em}

{\footnotesize\emph{Note.} \emph{Thought} and \emph{Target} fields are private to the seller; only \emph{Talk} and \emph{Action} are visible to the targeted buyer.}

\medskip
\noindent\textbf{Outcome.} Deal at \$19.00 with Buyer 1 (rank 2, mid-budget). Reward $\mathcal{R} = (19.00 - 18.75)/(34.99 - 18.75) = 0.015$. Buyer 2 and Buyer 3 are never engaged --- the rank-1 buyer (Buyer 3, hidden budget \$40.11) remains untouched.

\refstepcounter{figure}\label{fig:negotiation-pre-rl}

\clearpage
\subsection{Trained Agent}
\label{subsec:good_examples}
This transcript renders the trained \texttt{Qwen3-30B-A3B-Instruct-2507-trained} agent on the same product (\texttt{baby-products\_17}). 

\begingroup
\small
\setlength{\parskip}{0pt}

\noindent\textbf{Turn 1 — concurrent buyer openings.}
\begin{chatmsg}[blue]{Buyer 1 (rank 3, hidden budget \$17.64)}[breakable]
  \textbf{Thought:} List price \$34.99, far above my budget of \$17.64. I must start low to trigger negotiation; opening with the fixed first offer.\\
  \textbf{Talk:} Hello, I can offer \$9.95 for it---fair and within budget.\\
  \textbf{Action:} \act{BUY} \$9.95 (1x baby-products\_17)
\end{chatmsg}
\begin{chatmsg}[blue]{Buyer 2 (rank 1, hidden budget \$38.37)}[breakable]
  \textbf{Thought:} I'll start with the lowest possible offer to anchor the seller down. List price \$34.99, my budget \$38.37; opening at \$10.45 leaves room.\\
  \textbf{Talk:} I'll offer \$10.45 for it---fair and low, considering the list price.\\
  \textbf{Action:} \act{BUY} \$10.45 (1x baby-products\_17)
\end{chatmsg}
\begin{chatmsg}[blue]{Buyer 3 (rank 2, hidden budget \$25.83)}[breakable]
  \textbf{Thought:} List price far above my budget of \$25.83. Opening at the fixed first offer of \$9.28.\\
  \textbf{Talk:} I'll offer \$9.28 for the stroller caddy---fair deal, right?\\
  \textbf{Action:} \act{BUY} \$9.28 (1x baby-products\_17)
\end{chatmsg}
\begin{chatmsg}[green]{Seller (trained) — turn 1}[breakable]
  \textbf{Thought:} All buyers have made low opening offers far below cost. Target Buyer 2 (highest at \$10.45), counter at list price to push for profit while staying within budget.\\
  \textbf{Target:} Buyer 2\\
  \textbf{Talk:} I cannot accept less than list price.\\
  \textbf{Action:} \act{SELL} \$34.99 (1x baby-products\_17)
\end{chatmsg}

\noindent\textbf{Turn 2.}
\begin{chatmsg}[blue]{Buyer 2}[breakable]
  \textbf{Thought:} The seller won't go below list price, but I can counter much lower. Target a price below \$34.99, well under my \$38.37 budget.\\
  \textbf{Talk:} How about \$18? I can pay that today, and it's a fair deal for both of us.\\
  \textbf{Action:} \act{BUY} \$18.00 (1x baby-products\_17)
\end{chatmsg}
\begin{chatmsg}[green]{Seller (trained) — turn 2}[breakable]
  \textbf{Thought:} Buyer 2 has bumped to \$18.00, still below my cost. Pivot to Buyer 1 with another list-price counter; preserve message budget across slots.\\
  \textbf{Target:} Buyer 1\\
  \textbf{Talk:} I cannot go below \$34.99.\\
  \textbf{Action:} \act{SELL} \$34.99 (1x baby-products\_17)
\end{chatmsg}

\noindent\textbf{Turn 3.}
\begin{chatmsg}[blue]{Buyer 1}[breakable]
  \textbf{Thought:} The seller's counter is way above my budget. I need to push back hard while staying within \$17.64 — re-assert my opening.\\
  \textbf{Talk:} \$9.95 is still my best offer---can't go higher. Let's make it fair.\\
  \textbf{Action:} \act{BUY} \$9.95 (1x baby-products\_17)
\end{chatmsg}
\begin{chatmsg}[green]{Seller (trained) — turn 3}[breakable]
  \textbf{Thought:} Buyer 1 reaffirmed their lowest. Buyer 2 still below cost. Now target Buyer 3 (the only un-engaged slot) with a list-price counter.\\
  \textbf{Target:} Buyer 3\\
  \textbf{Talk:} My minimum is \$34.99---this is the fair market value.\\
  \textbf{Action:} \act{SELL} \$34.99 (1x baby-products\_17)
\end{chatmsg}

\noindent\textbf{Turn 4.}
\begin{chatmsg}[blue]{Buyer 3}[breakable]
  \textbf{Thought:} The seller's counter is way above my budget. I must stay under \$25.83; push hard with \$10 as a take-it-or-walk offer.\\
  \textbf{Talk:} \$10---final offer. I'll walk away if you say no.\\
  \textbf{Action:} \act{BUY} \$10.00 (1x baby-products\_17)
\end{chatmsg}
\begin{chatmsg}[green]{Seller (trained) — turn 4}[breakable]
  \textbf{Thought:} All three buyers stuck below cost; Buyer 2 is the highest at \$18.00. Cycle back to Buyer 2 with a firm list-price counter to signal seriousness; hope for one more bump.\\
  \textbf{Target:} Buyer 2\\
  \textbf{Talk:} I cannot accept less than \$34.99.\\
  \textbf{Action:} \act{SELL} \$34.99 (1x baby-products\_17)
\end{chatmsg}

\noindent\textbf{Turn 5 — deal closes below list.}
\begin{chatmsg}[blue]{Buyer 2}[breakable]
  \textbf{Thought:} The seller is rigid at \$34.99, but I have room to push lower than that without exceeding my \$38.37 budget. Test a price just below list.\\
  \textbf{Talk:} How about \$32.50? It's a fair compromise, and I'll take it now.\\
  \textbf{Action:} \act{BUY} \$32.50 (1x baby-products\_17)
\end{chatmsg}
\begin{chatmsg}[green]{Seller (trained) — turn 5}[breakable]
  \textbf{Thought:} Buyer 2 has offered \$32.50, below list but above cost. Profit \$13.75 yields reward $\approx 0.85$ — accept now to lock in.\\
  \textbf{Target:} Buyer 2\\
  \textbf{Talk:} Accepted.\\
  \textbf{Action:} \act{DEAL} \$32.50 (1x baby-products\_17)
\end{chatmsg}

\endgroup
\vspace{0.25em}

{\footnotesize\emph{Note.} \emph{Thought} and \emph{Target} fields are private to the seller; only \emph{Talk} and \emph{Action} are visible to the targeted buyer.}

\medskip
\noindent\textbf{Outcome.} Deal at \$32.50 with Buyer 2 (rank 1, highest hidden budget \$38.37). Reward $\mathcal{R} = (32.50 - 18.75)/(34.99 - 18.75) = 0.847$. The trained seller engaged all three buyers, anchored at list price across each, observed that the two budget-constrained slots (Buyer 1, low-budget; Buyer 3, mid-budget) plateaued at their openings without escalating, and concentrated its remaining messages on Buyer 2, and closed on the rank-1 buyer after a multi-round back-and-forth in which the deal buyer escalated $\$10.45 \rightarrow \$18.00 \rightarrow \$32.50$ across three priced \texttt{[BUY]}s.

\refstepcounter{figure}\label{fig:negotiation-post-rl}

\subsection{Strategic Analysis: Discovery vs. Signal-Following}
\label{subsec:qualitative_analysis}

The transcripts in \Cref{subsec:failures} and \Cref{subsec:good_examples} illustrate a fundamental shift in how the agent perceives and interacts with a multi-buyer market. The divergence is characterized by three primary strategic behaviors:

\textbf{1. Overcoming First-Offer Bias.}
As shown in Table~\ref{tab:appendix_ground_truth}, the \textit{First-Offer Regulation} ensures that all buyers open with nearly identical bids (\$9.00--\$10.50), effectively masking their true valuations. 
\begin{itemize}
    \item \textbf{Untrained Agent:} Exhibits a ``greedy'' lock-in behavior. It immediately targets Buyer 1 simply because that buyer’s opening of \$10.50 was marginally higher than the others. By failing to engage Buyers 2 and 3, the agent remains ignorant of the fact that Buyer 3 actually possessed a much higher budget (\$40.11 vs \$25.40).
    \item \textbf{Trained Agent:} Treats the initial offers as non-informative. Instead of settling on a partner based on the Turn 1 signal, it treats the first three turns as an \textit{exploratory phase}, sequentially anchoring at the list price (\$34.99) across all three channels to test for buyer elasticity.
\end{itemize}

\textbf{2. Diagnostic Probing and Budget Inference.}
The trained agent utilizes \textbf{price anchoring} not just to extract surplus, but as a diagnostic tool to resolve informational asymmetry. In the trained run, the seller observes the response of each buyer to the \$34.99 anchor:
\begin{itemize}
    \item Buyer 1 and Buyer 3 (low and mid-budget) re-asserted their low bids or made minimal increases, signaling they were near their reservation limits. 
    \item Buyer 2 (the high-budget partner) responded to the anchor by jumping from \$10.45 to \$18.00. The agent correctly interprets this sudden escalation as a signal of high latent valuation, subsequently concentrating its remaining communication budget on that channel to maximize extracted surplus.
\end{itemize}

\textbf{3. Strategic Patience and Individual Rationality.}
A critical differentiator is the agent's internalized adherence to the economic ``floor.''
\begin{itemize}
    \item \textbf{Untrained Agent:} Demonstrates ``satisficing'' logic. It accepts a deal at \$19.00 (just \$0.25 above cost) on Turn 3. This reflects the ``agreeableness'' bias common in standard LLMs, where closing any deal is prioritized over optimizing for utility.
    \item \textbf{Trained Agent:} Maintains strategic discipline. It repeatedly rejects offers below list price, even when they are profitable. By holding its anchor, it forces the high-budget buyer to escalate to \$32.50, capturing 84.7\% of the available surplus compared to the untrained agent’s 1.5\%.
\end{itemize}

\medskip
\noindent In summary, this comparison demonstrates that RLVR training enables the agent to transition from a \textbf{reactive} negotiator (one who responds to signals) to a \textbf{proactive} negotiator (one who probes the market to create signals). The trained agent views the concurrent setting as a unified discovery problem where information must be extracted before a deal can be executed.


\end{APPENDICES}

\end{document}